\documentclass[11pt]{article}

\usepackage{graphicx,amsmath,amsfonts,amssymb,amsthm,subfigure,fullpage,url,setspace}

\newtheorem{proposition}{Proposition}
\newtheorem{corollary}{Corollary}

\newcommand{\g}{\,|\,}
\newcommand{\zap}[1]{}

\begin{document}

\title{\textbf{Distance Dependent Infinite Latent Feature Models}}

\author{Samuel J. Gershman$^1$, Peter I. Frazier$^2$ and David M. Blei$^3$ \\
$^1$ Department of Psychology and Princeton Neuroscience Institute, \\ Princeton University \\
$^2$ School of Operations Research and Information Engineering, \\ Cornell University \\
$^3$ Department of Computer Science, \\ Princeton University}

\date{}

\maketitle

\begin{abstract}
  Latent feature models are widely used to decompose data into a small
  number of components. Bayesian nonparametric variants of these
  models, which use the Indian buffet process (IBP) as a prior over
  latent features, allow the number of features to be determined from
  the data.  We present a generalization of the IBP, the
  \emph{distance dependent Indian buffet process} (dd-IBP), for
  modeling non-exchangeable data.  It relies on distances
  defined between data points, biasing nearby data to share more
  features.  The choice of distance measure allows for many kinds of
  dependencies, including temporal and spatial.  Further, the original
  IBP is a special case of the dd-IBP.  In this paper, we develop the
  dd-IBP and theoretically characterize its feature-sharing properties.  We derive a Markov chain Monte
  Carlo sampler for a linear Gaussian model with a dd-IBP prior and
  study its performance on several non-exchangeable data sets.
\end{abstract}

\newpage

\section{Introduction}

Many natural phenomena decompose into latent features.  For example,
visual scenes can be decomposed into objects; genetic regulatory
networks can be decomposed into transcription factors; music can be
decomposed into spectral components.  In these examples, multiple
latent features can be simultaneously active, and each can influence
the observed data. Dimensionality reduction methods, such as principal
component analysis, factor analysis, and probabilistic matrix
factorization, provide a statistical approach to inferring latent features \cite{bishop06}. These methods characterize a small set of dimensions, or features, and model each data point as a weighted combination of
these features. Dimensionality
reduction can improve predictions and identify hidden structures in observed data.

Dimensionality reduction methods typically require that the number of
latent features (i.e., the number of dimensions) be fixed in advance.
Researchers have recently proposed a more flexible approach based on
Bayesian nonparametric models, where the number of features is
inferred from the data through a posterior distribution.  These models
are usually based on the Indian buffet process
(IBP; \cite{griffiths05,griffiths11}), a prior over binary matrices
with a finite number of rows (corresponding to data points) and an
infinite number of columns (corresponding to latent features). Using
the IBP as a building block, Bayesian nonparametric latent feature
models have been applied to several statistical problems (e.g., \cite{knowles07,meeds07,miller09,navarro08}). Since the number of features is effectively unbounded, these models are sometimes known as ``infinite'' latent feature models.

The IBP assumes that data are \emph{exchangeable}: permuting the order
of rows leaves the probability of a feature matrix unchanged.  This
assumption may be appropriate for some data sets, but for many others
we expect dependencies between data points and, consequently, between
their latent representations. As examples, the latent features
describing human motion are autocorrelated over time; the
latent features describing environmental risk factors are
autocorrelated over space.  In this paper, we present a generalization
of the IBP---the \emph{distance dependent IBP} (dd-IBP)---that
addresses this limitation.  The dd-IBP allows infinite latent feature
models to capture non-exchangeable structure.

The problem of adapting nonparametric models to non-exchangeable data
has been studied extensively in the mixture-modeling literature. In
particular, variants of the Dirichlet process mixture model allow dependencies between data points
(e.g., \cite{chung11,griffin06,caron07,duan07,rao09,ahmed10}). These dependencies may be spatial, temporal or more generally covariate-dependent; the effect of such dependencies is to induce sharing of features between nearby data points.

Among these methods is the \textit{distance dependent Chinese restaurant process}
(dd-CRP; \cite{blei10}).  The dd-CRP is a non-exchangeable
generalization of the Chinese restaurant process (CRP), the prior over
partitions of data that emerges in Bayesian nonparametric mixture
modeling \cite{blackwell73,escobar95,rasmussen00}. The dd-CRP models non-exchangeability by using using distances between
data points---nearby data points (e.g., in time or space) are more likely to be assigned to the same
mixture component.  The dd-IBP extends these ideas to infinite latent
feature models, where distances between data points influence feature-sharing, 
and nearby data points are more likely to share latent features.

We review the IBP in Section~\ref{sec:ibp}
and develop the dd-IBP in Section~\ref{sec:ddibp}. Like the dd-CRP, the dd-IBP lacks \emph{marginal invariance}, which means that removing one observation changes the distribution over the other observations. We discuss this property further in Section \ref{sec:invariance}. Although many Bayesian nonparametric models have this property, we view it as a particular modeling choice that may be appropriate for some problems but not for others.

Several other infinite latent feature models have been developed to
capture dependencies between data in different ways, for example using
phylogenetic trees \cite{miller08} or latent Gaussian processes
\cite{williamson10}. Of particular relevance to this work is the
model of Zhou et al. \cite{zhou11}, which uses a hierarchical beta process to couple data.  These and other related models are
discussed further in Section \ref{sec:related}. In Section
\ref{sec:feat}, we characterize the feature-sharing properties of the
dd-IBP and compare it to those of the model proposed by
Zhou et al. \cite{zhou11}. We find that the different models capture
qualitatively distinct dependency structures.

Exact posterior inference in the dd-IBP is intractable.  We present an
approximate inference algorithm based on Markov chain Monte Carlo
(MCMC; \cite{robert04}) in Section \ref{sec:mcmc}, and we apply
this algorithm in Section~\ref{sec:lin} to infer the latent features
in a linear-Gaussian model.  The experimental results presented in
Section \ref{sec:results} suggest that the dd-IBP is an effective tool
for modeling latent structure in data with dependencies between observations.

\section{The distance dependent Indian buffet process}

We first review the definition of the IBP and
its role in defining infinite latent feature models.  We then
introduce the dd-IBP.

\subsection{The Indian buffet process}
\label{sec:ibp}

The IBP is a prior over binary matrices $\mathbf{Z}$
with an infinite number of columns \cite{griffiths05,griffiths11}.
In the Indian buffet metaphor, rows of $\mathbf{Z}$ correspond to
customers and columns correspond to dishes.  In data analysis, the
customers represent data points and the dishes represent features. Let $z_{ik}$ denote the entry of $\mathbf{Z}$ at row $i$ and column $k$.
Whether customer $i$ has decided to sample dish $k$ (that is, whether
$z_{ik}=1$) corresponds to whether data point $i$ possesses feature
$k$.

The IBP is defined as a sequential process.  The first customer enters
the restaurant and samples the first $\lambda_1 \sim \mbox{Poisson}(\alpha)$ number of
dishes, where the hyperparameter $\alpha$ is a scalar.  In the binary
matrix, this corresponds to the first row being a contiguous block of
ones, whose length is the number of dishes sampled ($\lambda_1$), followed by an infinite block of zeros.

\begin{figure*}
\centering
\includegraphics[width=0.5\textwidth]{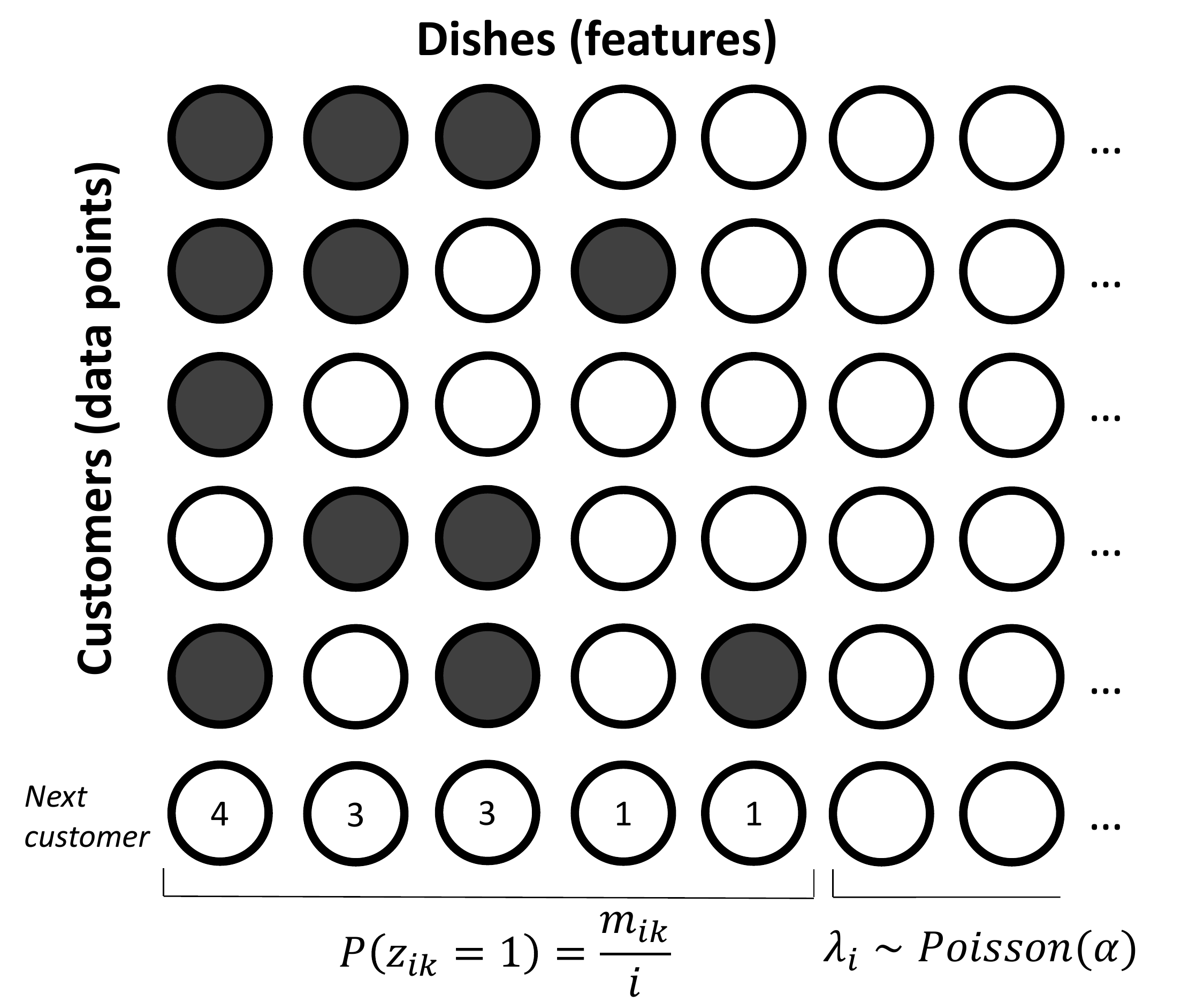}
\caption{\textbf{Schematic of the IBP}. An example of a latent
  feature matrix ($\mathbf{Z}$) generated by the IBP. Rows correspond to customers
  (data points) and columns correspond to dishes (features). Gray shading indicates that
  a feature is active for a given data point. The last row illustrates the assignment process for a new customer; the counts for each feature ($m_{ik}$) are shown inside the circles for previously sampled features.}
	\label{fig:ibp_schematic}
\end{figure*}

Subsequent customers $i=2,\ldots,N$ enter,
sampling each previously sampled dish according to its
popularity,
\begin{equation}
  p(z_{ik} = 1 \g \mathbf{z}_{1:(i-1)}) = m_{ik} / i,
  \label{eq:ibp}
\end{equation}
where $m_{ik} = \sum_{j \leq i} z_{jk}$ is the number of customers that sampled dish $k$ prior to customer $i$.
(We emphasize that Eq. \ref{eq:ibp} applies only to dishes $k$ that were previously
sampled, i.e., for which $m_{ik}>0$.)  Then, each customer samples $\lambda_i \sim \textrm{Poisson}(\alpha/i)$ new dishes. Again these are represented as a contiguous block of ones in the columns beyond the last dish sampled by a previous customer.

Though described sequentially, Griffiths and Ghahramani \cite{griffiths05} showed that
the resulting rows of the binary matrix are \textit{exchangeable} (up to a
permutation of the columns).  This means that the order of the
customers does not affect the probability of the resulting binary
matrix. This is seen in the Beta-Bernoulli perspective, which we
review in Section~\ref{sec:related}. In the next section, we develop
a generalization of the IBP that relaxes this assumption.

\subsection{The distance dependent Indian buffet process}
\label{sec:ddibp}

Like the IBP, the dd-IBP is
a distribution over binary latent feature matrices with a finite
number of rows and an infinite number of columns. Each pair of customers has an associated distance, e.g., distance in
time or space, or based on a covariate.  Two customers that are close
together in this distance will be more likely to share the same dishes
(that is, features) than two customers that are far apart.

The dd-IBP can be understood in terms of the following sequential
construction.  First, each customer selects a Poisson-distributed
number of dishes (feature columns). The dishes selected by a customer in this phase of the construction are said to be ``owned'' by this customer. A dish is either unowned, or is owned by exactly one customer. This step is akin to the selection of new dishes in the IBP.

Then, for each owned dish, customers connect to
one another.  The probability that one customer connects to another
decreases in the distance between them.  Note that customers do not
sample each dish, as in the IBP, but rather connect to other
customers.

Finally, dish inheritance is computed: A customer inherits a dish if its owner (from the first step)
is reachable in the connectivity graph for that dish. This inheritance is computed deterministically from the connections generated in the previous step.\footnote{If one insists
upon a complete gastronomical metaphor, customer connectivity can be
thought of as ``I'll have what he's having.''}  The dishes that each
customer samples are those that he inherits or owns.  Thus, 
similarity of sampled dishes between nearby
customers is induced via distance-dependent connection probabilities.

We now more formally describe the probabilistic generative process of
the binary matrix $\textbf{Z}$.  First, we introduce some notation and
terminology.
\begin{itemize}
\item Dishes (columns of $\mathbf{Z}$) are identified with the natural numbers $\mathbb{N} = \lbrace 1,2,\ldots \rbrace$. The set of dishes owned by customer $i$ is $\mathcal{K}_i \subset \mathbb{N}$. The cardinality of this
  set is $\lambda_i = |\mathcal{K}_i|$. These sets are disjoint, so $\mathcal{K}_i \cap \mathcal{K}_j = \emptyset$ for $i \neq j$. The total number of owned
  dishes is $K=\sum_{i=1}^N \lambda_i$. The set of dishes owned by customers excluding $i$ is
  $\mathcal{K}_{-i} = \cup_{j \neq i} \mathcal{K}_j$.

\item Each dish is associated with a set of customer-to-customer assignments, specified by the $N \times K$ \emph{connectivity matrix} $\mathbf{C}$, where
  $c_{ik}=j$ indicates that customer $i$ connects to customer $j$ for
  dish $k$.  Given $\mathbf{C}$, the customers form a set of (possibly
  cyclic) directed graphs, one for each dish. The \textit{ownership vector} is $\mathbf{c}^\ast$, where
  $c^\ast_k \in \lbrace 1,\ldots,N \rbrace$ indicates the customer who
  owns dish $k$, so $c^\ast_k = i \iff k \in \mathcal{K}_i$.

\item The $N \times N$
  distance matrix between customers is $\textbf{D}$, where the
  distance between customers $i$ and $j$ is $d_{ij}$. A customer's self-distance is 0: $d_{ii} =0$.
  We call the distance matrix \textit{sequential} when $d_{ij}=\infty$ for $j > i$. In this
  special case, customers can only connect to previous customers.  

\item The \textit{decay function} $f: \mathbb{R} \mapsto [0,1]$ maps distance to a quantity, which we call \textit{proximity}, that controls the probabilities of customer links.  We require that $f(0)=1$ and $f(\infty) = 0$. We obtain the \textit{normalized proximity matrix} $\mathbf{A}$ by
  applying the decay function to each customer pair and normalizing by
  customer. That is, $a_{ij} = f(d_{ij})/h_i$, where $h_i =
  \sum_{j=1}^N f(d_{ij})$.
\end{itemize}

Using this notation, we generate the feature indicator matrix
$\mathbf{Z}$ as follows:

\begin{enumerate}

\item \textbf{Assign dish ownership}. For each customer $i$, allocate $\lambda_i \sim \mbox{Poisson}(\alpha/h_i)$ unowned dishes to the customer's set of owned dishes, $\mathcal{K}_i$. For each $k \in \mathcal{K}_i$, set the ownership $c^\ast_k=i$.

\item \textbf{Assign customer connections}. For each customer $i$ and
  dish $k  \in \lbrace 1,\ldots,K \rbrace$, draw a customer assignment according
  to $P(c_{ik}=j|\mathbf{D},f) = a_{ij}, j = 1,\ldots,N$. Note that customers can
  connect to themselves. In this case, they do not inherit a dish
  unless they own it (see the next step).

\item \textbf{Compute dish inheritance}. We say that customer $j$
  \textit{inherits} dish $k$ if there exists a path along the directed
  graph for dish $k$ from customer $j$ to the dish's owner $c^\ast_k$ (i.e., if $c^\ast_k$ is reachable from $j$), where the directed graph is defined by column $k$ of $\mathbf{C}$. The owner of a dish automatically inherits it.\footnote{Although customer $i$ can link to other customers for dish $k$ even if $k \in \mathcal{K}_i$, these connections are ignored in determining dish inheritance when $k \in \mathcal{K}_i$.} We encode reachability with $\mathcal{L}$.  If customer $j$ is reachable from customer $i$ for dish $k$ then $\mathcal{L}_{ijk}=1$. Otherwise $\mathcal{L}_{ijk}=0$.

\item \textbf{Compute the feature indicator matrix}. For each customer
  $i$ and dish $k$ we set $z_{ik}=1$ if $i$ inherits $k$, otherwise $z_{ik}=0$.

\end{enumerate}

\begin{figure*}
\centering
\includegraphics[width=0.3\textwidth]{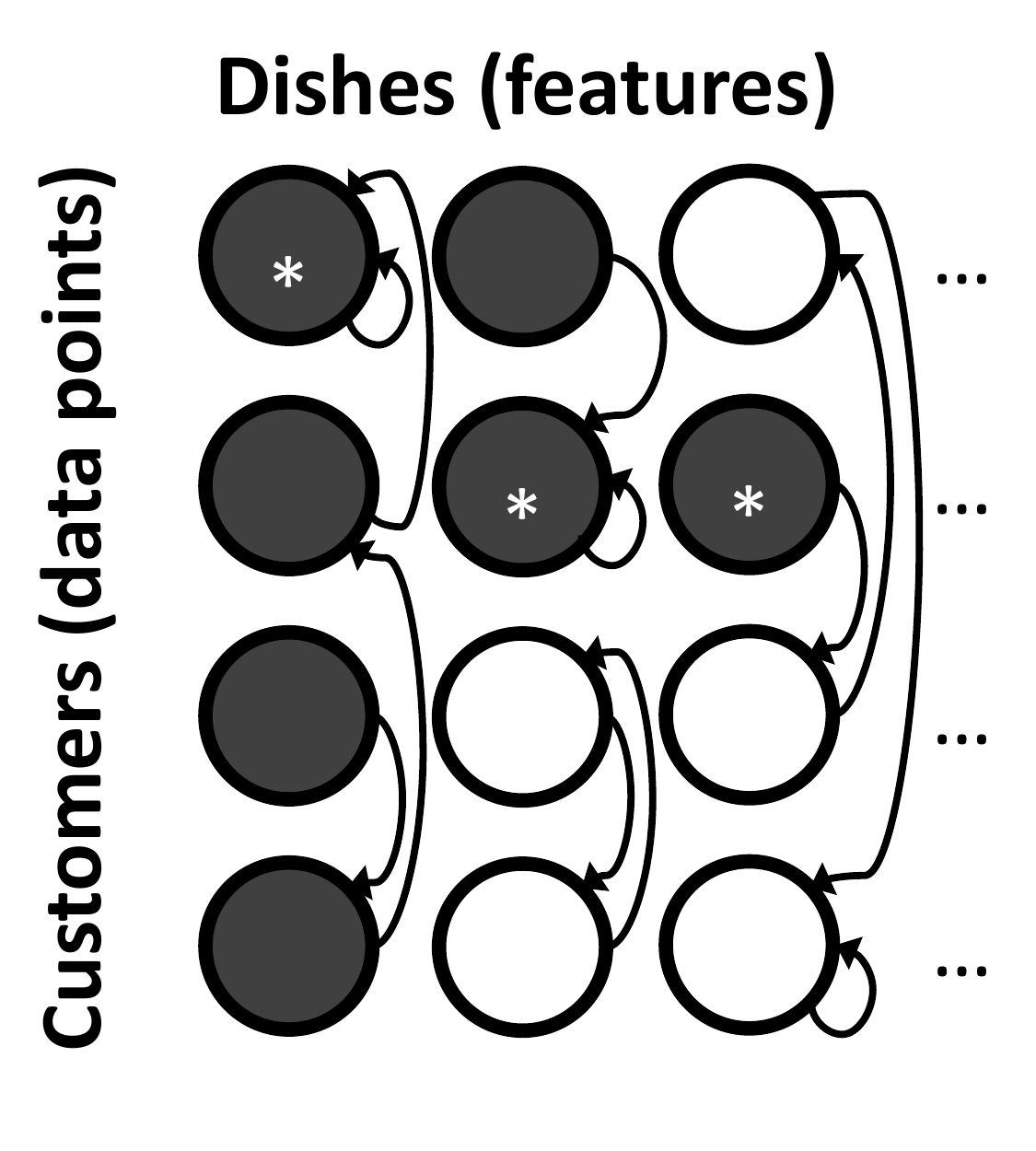}
\caption{\textbf{Schematic of the dd-IBP}. An example of a latent
  feature matrix generated by the dd-IBP. Rows correspond to customers
  (data points) and columns correspond to dishes (features). Customers
  connect to each other, as indicated by arrows. Customers inherit a
  dish if the owner of that dish ($c^\ast_k$, indicated by stars) is
  reachable by a sequence of connections. Gray shading indicates that
  a feature is active for a given data point.}
	\label{fig:ddibp_schematic}
\end{figure*}

An example of customer assignments sampled from the dd-IBP is shown in
Figure \ref{fig:ddibp_schematic}. In this example, customer 1 owns dish 1; customers 2-4 all reach customer 1, either directly or through a chain, and thereby inherit the dish (indicated by gray shading). Consequently, feature 1 is active for customers 1-4. Dish 2 is owned by customer 2; only customer 1 reaches customer 2, and hence feature 2 is active for customers 1 and 2. Dish 3 is owned by customer 2, but no other customers reach customer 2, and hence feature 3 is active only for that customer.

The generative process of the dd-IBP defines the following joint
distribution of the ownership vector and connectivity matrix,
\begin{equation}
  P(\textbf{C},\textbf{c}^* \g \mathbf{D}, \alpha, f) =
  P(\textbf{c}^* \g \alpha) P(\textbf{C} \g \textbf{c}^*, \mathbf{D}, f).
\end{equation}
Consider the first term.  Recall that the set of dishes each
customer owns $\mathcal{K}_i$ and the total number of owned dishes $K$
are both functions of the ownership vector $\textbf{c}^*$.  Thus, the
probability of the ownership vector is
\begin{equation}
  P(\textbf{c}^* \g \alpha) = \prod_{i=1}^{N}
  P(\lambda_i \g \alpha),
\end{equation}
where $\mathbf{c}^\ast$ is a deterministic function of $\lambda_1,\ldots,\lambda_N$.

Consider the second term.  The conditional distribution of the
connectivity matrix $\mathbf{C}$ depends on the total number of owned
dishes and the normalized proximity matrix $\mathbf{A}$ (derived from the
distances and decay function),
\begin{align}
  P(\mathbf{C}|\textbf{c}^*,\mathbf{D},f) = \prod_{i=1}^N \prod_{k=1}^K
  a_{ic_{ik}}.
\end{align}
The dependence on $\textbf{c}^*$ comes from $K$, which is determined by $\textbf{c}^*$.

Random feature models (and the traditional IBP)
operate with a random binary matrix $\textbf{Z}$.  In the dd-IBP,
$\textbf{Z}$ is a (deterministic) many-to-one function of the random variables $\mathbf{C}$ and $\mathbf{c}^\ast$, which
we denote by $\phi$.  We compute the probability of a binary matrix by
marginalizing out the appropriate configurations of these variables
\begin{equation}
  P(\mathbf{Z}|\mathbf{D},\alpha, f) = \sum_{(\mathbf{c}^\ast,
    \mathbf{C}): \phi(\mathbf{c}^\ast, \mathbf{C})=\mathbf{Z}}
    P(\mathbf{c}^\ast, \textbf{C}|\mathbf{D}, \alpha, f).
\end{equation}
The dd-IBP reduces to the standard IBP in the special case 
when $f(d)=1$ for all $d < \infty$ and the distance matrix is sequential. 
(Recall: $\mathbf{D}$ is sequential if $d_{ij}=\infty$ for $j>i$.)
To see this, consider the
probability that the $k$th dish is sampled by the $i$th customer (that
is, $z_{ik}=1$).  This probability is the proportion of previous
customers that already reach $c^\ast_k$ because the probabiity of
connecting to each customer is proportional to one. This probability is $m_{ki}/i$, which is the same as in the IBP. This is akin to the relationship between the dd-CRP and the traditional CRP under the same condition.

\begin{figure*}
\centering
\includegraphics[width=0.9\textwidth]{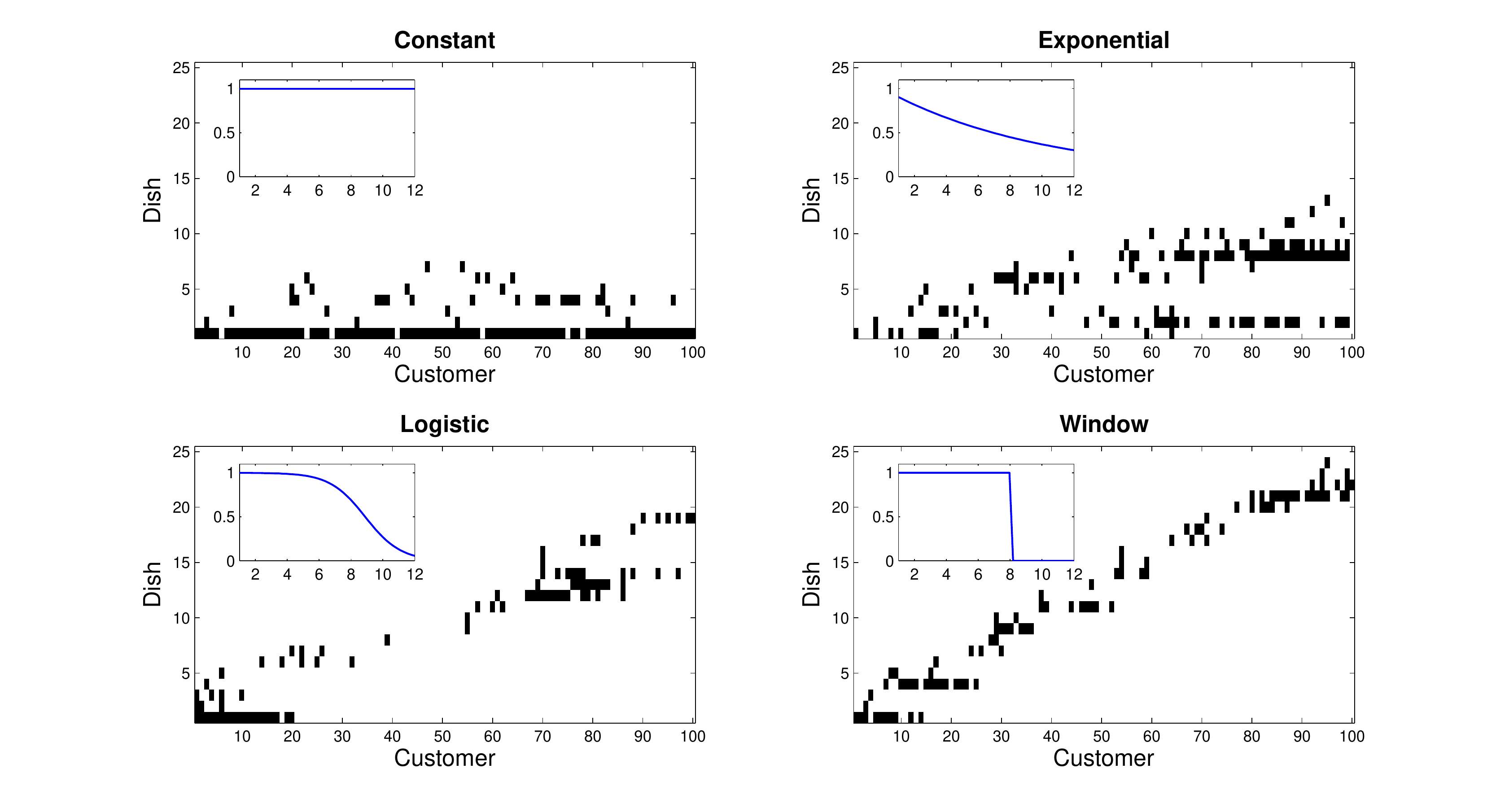}
  	\caption{\textbf{Decay functions}. Each panel presents a
          different latent feature matrix, sampled from the dd-IBP
          with sequential distances. Decay functions are shown in
          the insets.}
	\label{fig:decayfun}
\end{figure*}

Many different decay functions are possible within this
framework. Figure \ref{fig:decayfun} shows samples of $\mathbf{Z}$
using four decay functions and a sequential distance defined by absolute temporal distance ($d_{ij} = i-j$ for $i \geq j$ and $d_{ij} = \infty$ for $j>i$).
\begin{itemize}
\item The \emph{constant}, $f(d)=1$. This is the standard IBP.
\item The \emph{exponential}, $f(d) = \exp(-\beta d)$.
\item The \emph{logistic}, $f(d) = 1/(1+\exp(\beta d - \nu))$.
\item The \emph{window}, $f(d) = \mathbf{1}[d < \nu]$.
\end{itemize}
Each decay function encourages the sharing of features across nearby
rows in a different way.

When combined with an observation model (which specifies how the latent features give rise to observed data), the dd-IBP functions as a prior over latent feature representations of a data set. In section \ref{sec:lin}, we consider a specific example of how the dd-IBP can be used to analyze data.

\subsection{Marginal invariance and exchangeability}
\label{sec:invariance}

Unlike the traditional IBP, the dd-IBP is not (in general)
\emph{marginally invariant}, the property that removing a customer
leaves the distribution over latent features for the remaining
customers unchanged.  (The dd-IBP builds on the dd-CRP, which is not
marginally invariant either.) In some circumstances, marginal
invariance is desirable for computational reasons. For example, the
conditional distributions over missing data for models lacking
marginal invariance require computing ratios of normalization
constants. In contrast, marginally invariant models, due to their
factorized structure, require less computation for conditional
distributions over missing data.  In other circumstances, such as 
exploratory analysis of fully observed datasets, this computational concern is less important.   
Beyond computational considerations, while marginal invariance may be an
appropriate modeling assumption in some data sets, it may be inappropriate in others.  

Also unlike the traditional IBP, the dd-IBP is not exchangeable in general.
To state this formally, 
let $\pi$ be a permutation of the integers $\{1,\ldots,N\}$,
and for a given $N\times \infty$ binary matrix $\mathbf{Z}$, let $\mathbf{Z}^\pi$ be the matrix created by permuting its rows according to $\pi$.
Let $\mathbf{Z}$ be drawn from the dd-IBP with distance matrix $\mathbf{D}$, mass parameter $\alpha$ and decay function $f$.
Then, except in certain special cases 
(such as when $\mathbf{D}$ recovers the traditional IBP), 
\begin{equation*}
P(\mathbf{Z} = \mathbf{z}|\mathbf{D},\alpha,f) \ne P(\mathbf{Z} = \mathbf{Z}^\pi|\mathbf{D},\alpha,f).
\end{equation*}
Permuting the data changes its distribution, and so the dd-IBP is not exchangeable in general.

Although the dd-IBP is not exchangeable, it does have a related symmetry.
Let $\mathbf{D}^\pi$ be the $N\times N$ matrix $\mathbf{D}$ with both its rows and its columns permuted according to
$\pi$, and let $\mathbf{Z}^\pi$ be drawn from the dd-IBP with distance matrix $\mathbf{D}^\pi$ rather than $\mathbf{D}$.
(We retain the same values for $\alpha$ and $f$.)
Then, in general,
\begin{equation*}
P(\mathbf{Z} = \mathbf{z}|\mathbf{D},\alpha,f) = P(\mathbf{Z}^\pi = \mathbf{z}^\pi|\mathbf{D},\alpha,f).
\end{equation*}
Thus, if we permute both the data and the distance matrix, probabilities remain unchanged.  Permuting
both the data and the distance matrix is like first relabeling the data, and then explicitly altering the probability
distribution to account for this relabeling. If the dd-IBP were exchangeable, one would not need to alter the probability distribution to account for relabeling. 

\section{Related work}
\label{sec:related}

In this section we describe related work on infinite latent feature
models that capture external dependence between the data.  We 
focus on the most closely related model, which is the
\textit{dependent hierarchical beta process} (dHBP; \cite{zhou11}).
As a prelude to describing the dHBP, we review the connection between
the IBP and the beta process.

\subsection{The beta process}

Recall that the IBP is exchangeable. Consequently, by de Finetti's
theorem \cite{bernardo94}, the rows of $\mathbf{Z}$, considered as binary vectors $\lbrace \mathbf{z}_i \rbrace$, are conditionally
independent,
\begin{align}
  P(\mathbf{Z}) = \int \prod_{i=1}^N P(\mathbf{z}_i|B) dP(B).
\end{align}
In this marginal distribution, $B$ is a random measure on the feature space
$\Omega$ and $P(B)$ is the de Finetti mixing distribution
(see \cite{bernardo94}). Thibaux and Jordan \cite{thibaux07} showed that the de
Finetti mixing distribution underlying the IBP is the \emph{beta
  process} (BP), parameterized by a positive \emph{concentration
  parameter} $c$ and a \emph{base measure} $B_0$ on $\Omega$. A draw
$B \sim \mbox{BP}(c,B_0)$ is defined by a countably infinite
collection of weighted atoms,
\begin{align}
B = \sum_{k=1}^\infty p_k \delta_{\omega_k},
\end{align}
where $\delta_{\omega}$ is a probability distribution that places a single atom at $\omega \in \Omega$, and the $p_k \in [0,1]$ are independent random variables whose distribution is described as follows. If $B_0$ is continuous, then the atoms and their weights are drawn from a nonhomogeneous Poisson process defined on the space $\Omega \times [0,1]$ with rate measure
\begin{align}
\nu(d\omega,dp) = c p^{-1} (1-p)^{c-1} dp B_0(d\omega).
\end{align}
If $B_0$ is discrete and of the form $B_0 = \sum_{k=1}^\infty q_k
\delta_{\omega_k}$, $q_k \in [0,1]$, then $B$ has atoms at the same locations as $B_0$, with $p_k \sim \mbox{Beta}(c q_k,c
(1-q_k))$. Following Thibaux and Jordan \cite{thibaux07}, we define the \emph{mass
  parameter} as $\gamma = B_0(\Omega)$. Note that $B_0$ is not necessarily a
probability measure, and hence $\gamma$ can take on non-negative values different from 1.

Conditional on a draw from the beta process, the feature
representation $X_i$ of data point $i$ is generated by drawing from the
\emph{Bernoulli process} (BeP) with base measure $B$: $X_i|B \sim
\mbox{BeP}(B)$. 
If $B$ is discrete, then $X_i = \sum_{k=1}^\infty z_{ik} \delta_{\omega_k}$, where $z_{ik} \sim \mbox{Bernoulli}(p_k)$. In other words, feature $k$ is activated with probability $p_k$ independently for all data points. Sampling $\mathbf{Z}$ from the compound
beta-Bernoulli process is equivalent to sampling $\mathbf{Z}$ directly
from the IBP when $c=1$ and $\gamma = \alpha$ \cite{thibaux07}.

\subsection{Dependent hierarchical beta processes}
\label{s:dHBP}

The dHBP \cite{zhou11} builds
external dependence between data points using the above BP construction.
The dependencies are induced by mixing
independent BP random measures, weighted by their proximities
$\mathbf{A}$.

The dHBP is based on the following generative process,
\begin{align}
  &X_i|B^\ast_{g_i} \sim \mbox{BeP}(B^\ast_{g_i}), \quad g_i \sim
  \mbox{Multinomial}(\mathbf{a}_i), \nonumber \\
  &B^\ast_j|B \sim \mbox{BP}(c_1,B), \quad B \sim \mbox{BP}(c_0,B_0).
\end{align}
This is equivalent to drawing $X_i$ from a Bernoulli process whose
base measure is a linear combination of BP random measures,
\begin{align}
X_i|B_i \sim \mbox{BeP}(B_i), \quad B_i = \sum_{j=1}^N a_{ij} B^\ast_j.
\end{align}
Dependencies between data points are captured in the dHBP by the
proximity matrix $\mathbf{A}$, as in the
dd-IBP.\footnote{Zhou et al. \cite{zhou11} formalize dependencies in an
  equivalent manner using a normalized kernel function defined over
  pairs of covariates associated with the data points.} This allows
proximal data points (e.g., in time or space) to share more latent
features than distant ones.

In Section \ref{sec:feat}, we compare the feature-sharing
properties of the dHBP and dd-IBP.
Using an asymptotic analysis, we show that the
dd-IBP offers more flexibility in modeling the proportion of features shared between data points, but less flexibility in
modeling uncertainty about these proportions.

\subsection{Other non-exchangeable variants}

Although still a nascent area of research, several other
non-exchangeable priors for infinite latent feature models have been
proposed. Williamson, Orbanz and Ghahramani \cite{williamson10} used a hierarchical Gaussian process to
couple the latent features of data in a covariate-dependent manner. They named this model the \emph{dependent Indian buffet process} (dIBP).
Their framework is flexible: It can couple columns of
$\mathbf{Z}$ in addition to rows, while the dd-IBP cannot. However,
this flexibility comes at a computational cost during inference: Their
algorithm requires sampling an extra layer of variables.

Miller, Griffiths and Jordan \cite{miller08} proposed a ``phylogenetic IBP'' that encodes
tree-structured dependencies between data. Doshi-Velez and Ghahramani \cite{doshi09b} proposed a
``correlated IBP'' that couples data points and features through a set
of latent clusters. Both of these models relax exchangeability, but
they do not allow dependencies to be specified directly in terms of
distances between data.  Furthermore, inference for these models
requires more intensive computation than does the standard IBP. The
MCMC algorithm presented by Miller et al. \cite{miller08} for the phylogenetic IBP
involves both dynamic programming and auxiliary variable
sampling. Similarly, the MCMC algorithm for the correlated IBP
involves sampling latent clusters in addition to latent features. Our model also incurs extra computational cost relative to the traditional IBP due to the computation of reachability (quadratic in the number of observations); however, it permits a richer specification of the dependency structure between observations than either the phylogenetic or the correlated IBP.

Recently, Ren et al. \cite{ren11} presented a novel way of introducing dependency into latent feature models based on the beta process. Instead of defining distances between customers, each dish is associated with a latent covariate vector, and distances are defined between each customer's (observed) covariates and the dish-specific covariates. Customers then choose dishes with probability proportional to the customer-dish proximity. This construction comes with a significant computational advantage for data sets where the time complexity is tied predominantly to the number of observations. The downside of this construction is that the MCMC algorithm used for inference must sample a separate covariate vector for each dish, which may scale poorly if the covariate dimensionality is high.

\section{Characterizing feature-sharing}
\label{sec:feat}

In this section, we compare the feature-sharing properties of the dHBP and dd-IBP. 
Two data points share a feature if that feature is active for both (i.e., $z_{ik}=z_{jk}=1$ for $i \neq j$ and a given feature $k$). This analysis is useful for understanding the types of dependencies induced by the different models, and can help guide the choice of model and hyperparameter settings for particular data analysis problems. We consider an asymptotic regime in which the mass parameter is large ($\alpha$ for the dd-IBP and $\gamma$ for the dHBP), which simplifies feature-sharing properties.  Proofs of all propositions in this section may be found in the Appendix.

\subsection{Feature-sharing in the dd-IBP}
\label{sec:ddibp-theory}

We first characterize the limiting distributional properties of
feature-sharing in the dd-IBP as $\alpha \rightarrow \infty$. We drop
the feature index $k$ in the reachability indicator
$\mathcal{L}_{ijk}$, writing it $\mathcal{L}_{ij}$.  We do this
because features (that is, columns of the binary matrix $\mathbf{Z}$) are
exchangeable under the dd-IBP and, consequently, the distribution of
the random vector $(\mathcal{L}_{ijk} : i,j=1,\ldots,n)$ is invariant
across $k$.

Let $R_i = \sum_{k=1}^\infty z_{ik}$ denote the number of features held by data point $i$, and let $R_{ij}=\sum_{k=1}^\infty z_{ik}z_{jk}$
  denote the number of features shared by data points $i$ and $j$, where $i\ne j$.
\begin{proposition}
  Under the dd-IBP,
\begin{align}
R_i &\sim \mathrm{Poisson}\left( \alpha \sum_{n=1}^N h_n^{-1} P(\mathcal{L}_{in}=1) \right), \\
R_{ij} &\sim \mathrm{Poisson} \left( \alpha \sum_{n=1}^N h_n^{-1} P(\mathcal{L}_{in}=1,\mathcal{L}_{jn}=1) \right).
\end{align}
\end{proposition}
The probabilities $P(\mathcal{L}_{in}=1)$ and $P(\mathcal{L}_{in}=1,\mathcal{L}_{jn}=1)$ depend strongly on the distribution of the connectivity matrix $\mathbf{C}$, but do not depend on the ownership vector $\mathbf{c}^\ast$, since $\mathcal{L}$ is independent of dish ownership.

We derive the limiting properties of $R_i$ and $R_{ij}$ from
properties of the Poisson distribution.  In this and following
results, $\xrightarrow{d}$ indicates convergence in distribution.
\begin{corollary}
  \label{cor:ddibp}
  Let $i\ne j$.
  $R_i$ and $R_{ij}$ converge in distribution under the dd-IBP to the
  following constants as $\alpha \rightarrow \infty$:
\begin{align}
\frac{R_i}{\alpha} &\xrightarrow{d} \sum_{n=1}^N h_n^{-1} P(\mathcal{L}_{in}=1), \\
\frac{R_{ij}}{\alpha} &\xrightarrow{d} \sum_{n=1}^N h_n^{-1} P(\mathcal{L}_{in}=1,\mathcal{L}_{jn}=1), \\
\frac{R_{ij}}{R_i} &\xrightarrow{d} \frac{\sum_{n=1}^N h_n^{-1} P(\mathcal{L}_{in}=1,\mathcal{L}_{jn}=1)}{\sum_{n=1}^N h_n^{-1}P(\mathcal{L}_{in}=1)}.
\end{align}
\end{corollary}

This corollary shows that the limiting fraction of shared features
$R_{ij}/R_i$ in the dd-IBP is a constant that may be different for
each pair of data points $i$ and $j$.  
In contrast, we show below that
the same limiting fraction under the dHBP is random, and takes one of
two values.  These two values are fixed, and do not depend upon the
data points $i$ and $j$.

\subsection{Feature-sharing in the dHBP}
\label{sec:dhbp-theory}

Here we characterize the limiting distributional properties of feature
sharing in the dHBP as $B_0$ becomes infinitely concentrated (i.e.,
$\gamma \rightarrow \infty$, analogous to $\alpha \rightarrow
\infty$). In this limit, feature-sharing is primarily attributable to
dependency induced by the proximity matrix $\mathbf{A}$.
\begin{proposition}
  If $B_0$ is continuous, then under the dHBP,
\begin{align}
  R_i|\mathbf{g}_{1:N} &\sim \mathrm{Poisson}\left(\gamma\right), \\
  R_{ij}|\mathbf{g}_{1:N} &\sim \left\{ \begin{array}{ll}
      \mathrm{Poisson}\left(\gamma\frac{c_0 + c_1 + 1}{(c_0+1)(c_1+1)} \right) & \mbox{if $g_i = g_j$},\\
      \mathrm{Poisson}\left(\gamma\frac{1}{c_0+1} \right) & \mbox{if
        $g_i \neq g_j$.}\end{array} \right.
\end{align}
\end{proposition}

We derive the limiting properties of $R_i$ and $R_{ij}$ from
properties of the Poisson distribution.
\begin{corollary}
  Let $i\ne j$.
  Conditional on $\mathbf{g}_{1:N}$, $R_i$ and $R_{ij}$ converge in
  distribution under the dHBP to the following constants as $\gamma
  \rightarrow \infty$:
\begin{align}
\frac{R_i}{\gamma}  &\xrightarrow{d} 1, \\
\frac{R_{ij}}{\gamma}
&\xrightarrow{d} \left\{ \begin{array}{ll}
      \frac{c_0 + c_1 + 1}{(c_0+1)(c_1+1)} & \mbox{if $g_i = g_j$},\\
      \frac{1}{c_0+1} & \mbox{if $g_i \neq g_j$,}\end{array} \right. \\
\frac{R_{ij}}{R_i}
&\xrightarrow{d} \left\{ \begin{array}{ll}
      \frac{c_0 + c_1 + 1}{(c_0+1)(c_1+1)} & \mbox{if $g_i = g_j$},\\
      \frac{1}{c_0+1} & \mbox{if $g_i \neq g_j$.}\end{array} \right.
\end{align}
\label{eq:R}
\end{corollary}
Thus, the expected fraction of object $i$'s features shared with
object $j$, $R_{ij}/R_i$, is a factor of $\frac{c_0 + c_1 + 1}{c_1+1}$
bigger when $g_i=g_j$. As $c_0 \rightarrow \infty$, this fraction goes
to $\infty$. As $c_0 \rightarrow 0$, it goes to 1. We can obtain the
unconditional fraction by marginalizing over $g_i$ and $g_j$:
\begin{corollary}
  \label{cor:dhbp}
  Let $i\ne j$.
$R_{ij}/R_i$ converges in distribution under the dHBP as $\gamma\rightarrow\infty$ to a random variable $M_{ij}$ defined by
\begin{equation}
  M_{ij} = 
  \begin{cases}
    \frac{c_0 + c_1 + 1}{(c_0+1)(c_1+1)} & \text{with probability $P(g_i=g_j)$}, \\
    \frac{1}{c_0+1} & \text{with probability $P(g_i\neq g_j)$},
  \end{cases}
\end{equation}
where $P(g_i = g_j) = \sum_{n=1}^N a_{in} a_{jn}$.
\end{corollary}
This corollary shows that as $\gamma$ grows large, the fraction of
shared features becomes one of two values (determined by $c_0$ and
$c_1$), with a mixing probability determined by the dependency
structure. Thus, the dHBP affords substantial flexibility in
specifying the mixing probability (via $\mathbf{A}$), but is
constrained to two possible values of the limiting fraction.

\subsection{Feature-sharing in the IBP}
\label{sec:ibp-theory}
For comparison, we briefly describe the feature-sharing properties under the traditional IBP.

Under the traditional IBP, by exchangeability, $R_i$ and $R_{ij}$ are equal in distribution to $R_1$ and $R_{12}$.
The first customer draws a $\mathrm{Poisson}(\alpha)$ number of dishes.
The second customer then chooses whether to sample each of these dishes independently and with probability $1/2$.
Thus, the number of dishes sampled by both the first and second customers is
$R_{12} \sim \mathrm{Poisson}(\alpha/2)$.

This shows that, under the traditional IBP, as $\alpha\rightarrow\infty$ with $i\ne j$,
\begin{equation}
  \frac{R_i}{\alpha} \xrightarrow{d} 1, \qquad
  \frac{R_{ij}}{\alpha} \xrightarrow{d} \frac12, \qquad
  \frac{R_{ij}}{R_i} \xrightarrow{d} \frac12.
\end{equation}

\subsection{Discussion}

Using an asymptotic analysis, the preceding theoretical results show that the
dd-IBP and dHBP provide different forms of flexibility in specifying the way in
which features are shared between data points.  This asymptotic analysis
takes the limit as the mass parameters $\alpha$ and $\gamma$ become large.  This
limit is taken for theoretical tractability, and removes much of the uncertainty
that is otherwise present in these models.  While such limiting dd-IBP and dHBP
models are not intended for practical use, their simplicity
provides insight into behavior in non-asymptotic regimes.

Under the dd-IBP, Corollary~\ref{cor:ddibp} shows that the modeler is allowed a
great deal of flexibility in specifying the proportions of features shared by
data points.  Given a matrix specifying the proportion of features that are
believed to be shared by pairs of data points, one can (if this matrix is
sufficiently well-behaved) design a distance matrix that causes the dd-IBP to
concentrate on the desired proportions.  While the dd-IBP cannot model an
arbitrary modeler-specified matrix of proportions, the set of matrices that can
be modeled is very large.

In contrast, under the dHBP, Corollary~\ref{cor:dhbp} shows that the modeler has
less flexibility in specifying the proportions of features shared.
Under the dHBP, the modeler chooses two values, $(c_0+c_1+1)/(c_0+1)(c_1+1)$ and
$1/(c_0+1)$, and the proportion of features shared by any pair of data points in
the asymptotic regime must be one of these two values. 

\begin{figure*}[tb]
\centering
\includegraphics[width=0.9\textwidth]{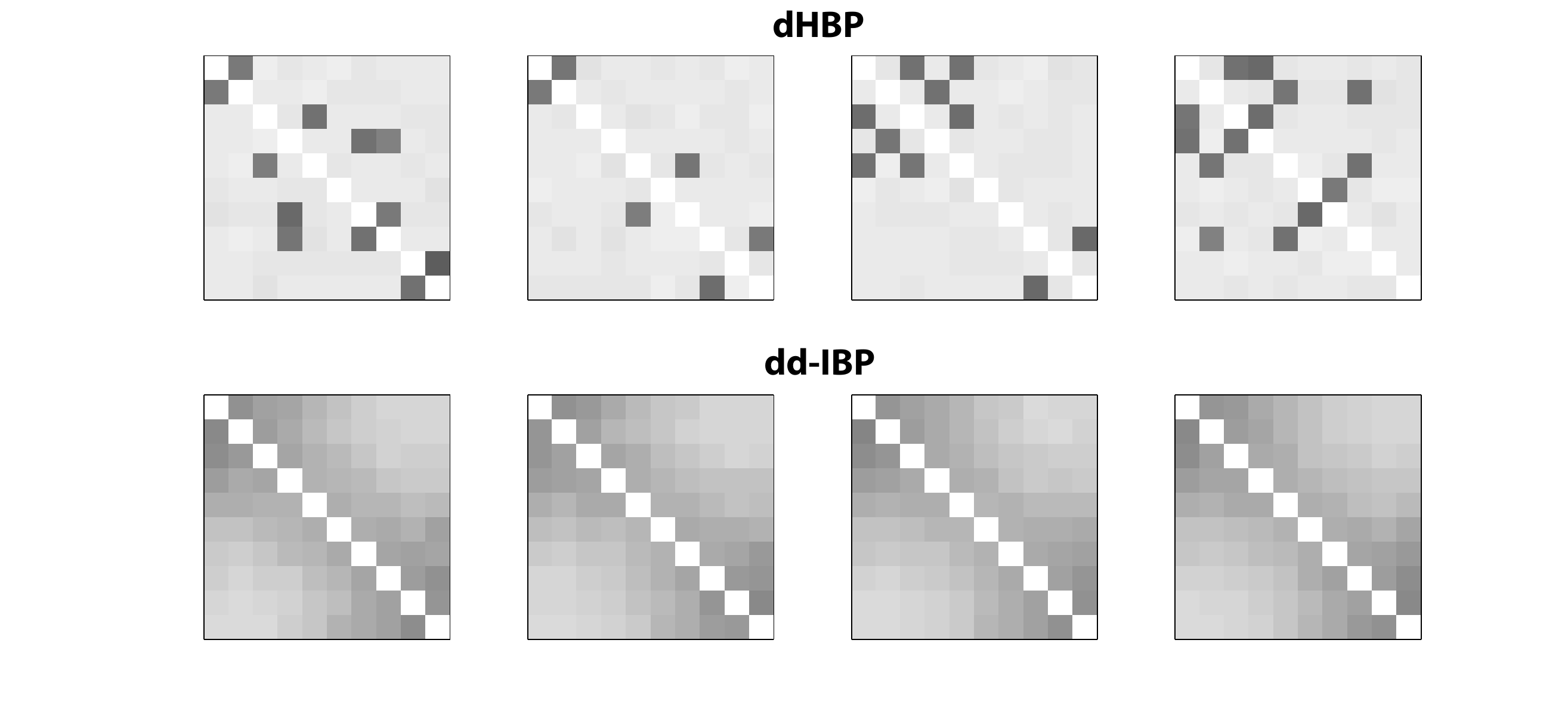}
\caption{\textbf{Feature-sharing in the dHBP and dd-IBP, limiting case}. 
  Along the horizontal axis, we show $4$ independent draws from the
  dHBP (\emph{Top}) and dd-IBP (\emph{Bottom}).
  Within each subfigure, the shade of a cell $(i,j)$ shows the fraction $R_{ij}/R_i$, where
  $R_i$ is the number of features held by data point $i$,
  and $R_{ij}$ is the number held by both $i$ and $j$.
  Diagonals $R_{ii}/R_i = 1$ have been set to $0$ for clarity.
  Here, $\alpha=\gamma=1000$.
  Limiting results from Section~\ref{sec:feat} explain the
  behavior for such large $\alpha$ and $\gamma$: for the dHBP the
  feature-sharing proportion $R_{ij}/R_i$ is random and equal to one of two constants;
  for the dd-IBP the proportion is non-random and takes a range of
  values.  
  The dd-IBP models feature-sharing proportions that differ across data points, 
  but does not model uncertainty about these proportions when mass parameters are large.
  \label{fig:featureshare}
  }
\end{figure*}

Section~\ref{sec:ibp-theory} shows that the traditional IBP has the least
flexibility.  In the asymptotic regime, the proportion of features shared by
each pair of data points is a constant. 

While the dd-IBP has more flexibility in specifying values of the
feature-sharing-proportions than the dHBP, it has less flexibility (at least in
this asymptotic regime) in modeling uncertainty about these feature-sharing
proportions.  Under the dd-IBP, the proportion of features shared by a pair of
data points in the asymptotic regime is a deterministic quantity.  Under the
dHBP, the proportion of features shared is a random quantity, even in the
asymptotic regime.  A modeler using the dHBP has full flexibility in choosing
the joint probability distribution governing these proportions.  One could
extend the dd-IBP to allow uncertainty about the feature-sharing-proportions by
specifying a hyperprior over distance matrices, but we do not consider this
extension further.

Figure~\ref{fig:featureshare} illustrates the difference in asymptotic
feature-sharing behavior between the dHBP and dd-IBP.  Subfigures in the upper
row are draws from the dHBP, and subfigures in the bottom row are draws from the
dd-IBP.
Within a single subfigure, the shade in the cell $(i,j)$ is the fraction $R_{ij}/R_i$.
(The diagonals $R_{ii}/R_i=1$ have been set to $0$ to bring out other aspects of the matrix.) 
Each of the four columns represents a pair of independent draws.  To approximate the asymptotic regime considered by the theory, the
mass parameters for the two models are set to large values of $\gamma=\alpha=1000$.
The figure shows that, in draws from the dHBP, off-diagonal cells have one of two shades, corresponding to the two
possible limiting values for $R_{ij}/R_i$.  In the different columns, correspoindng to different independent draws, the
patterns are different, showing that $R_{ij}/R_i$ remains random under the dHBP, even in the
asymptotic regime.  In contrast, in draws from the dd-IBP, off-diagonal cells take a variety of different values,
but remain unchanged across independent draws.

\begin{figure*}[tb]
\centering
\includegraphics[width=0.9\textwidth]{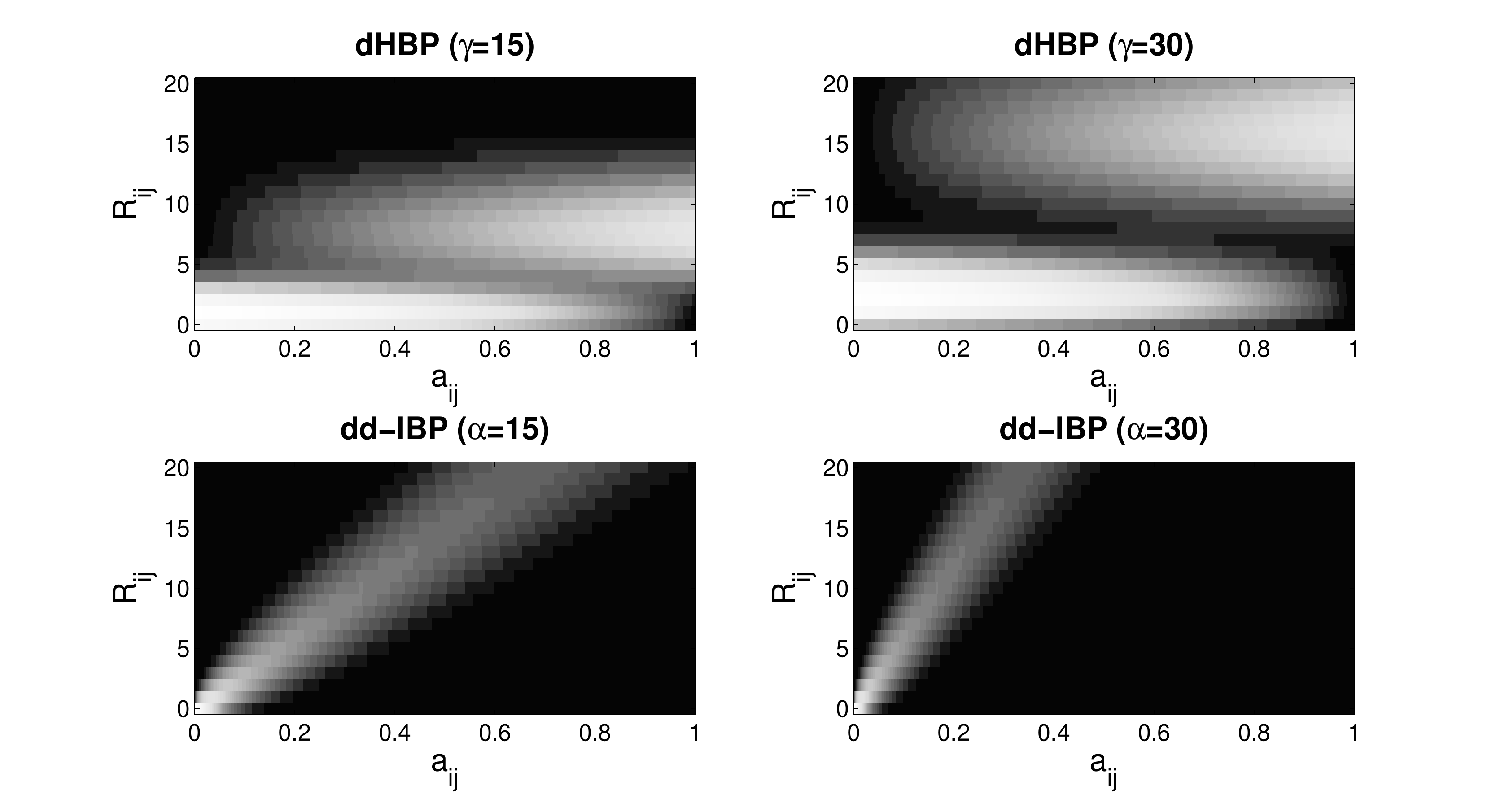}
\caption{\textbf{Feature-sharing in the dHBP and dd-IBP}. Heatmaps of
  the probability mass function over the number of shared features
  $R_{ij}$ (y-axis) as a function of proximity $a_{ij}$ (x-axis) in a data set consisting
  of two data points. Black indicates a probability mass of 0, with lighter shades indicating larger values. For the dHBP, we set $c_0=10$ and $c_1=1$. Note
  that $\mathbb{E}[R_i]$ is the same for both the dHBP and dd-IBP in
  these examples (when $\alpha=\gamma$).}
	\label{fig:featureshare_poisson}
\end{figure*}

Figure~\ref{fig:featureshare_poisson} illustrates non-asymptotic feature-sharing
behavior in a simple setting with only two data points.  The figure shows the feature-sharing behavior of the dHBP (top) and dd-IBP
(bottom) at two values for the mass parameter:
$\alpha=\gamma=15$ (top row) and $\alpha=\gamma=30$ (bottom row).  
Each subfigure shows the probability mass function
$P(R_{ij})$ as a function  the proximity $a_{ij}$, where $a_{ij} = 1/d_{ij}$ for the dd-IBP.
Because there are only two data points, with $a_{ii}=1$ and $a_{ij}=a_{ji}$,
specifying $a_{ij}$ is sufficient for specifying the full proximity matrix
$\mathbf{A}$.  For the dHBP, we set $c_0=10$ and $c_1=1$.  
Also facilitating comparison, $\mathbb{E}[R_i]$ is the same between both models (when $\alpha=\gamma$).

Figure~\ref{fig:featureshare_poisson} shows that as the proximity $a_{ij}$
increases to $1$, the number of shared features $R_{ij}$ tends to increase under
both models.  More precisely, $P(R_{ij})$ 
concentrates on larger values of $R_{ij}$ as $a_{ij}$ increases.  However, the way in
which the probability mass functions change with $a_{ij}$ differs 
between the two models.  In the dd-IBP, the most likely value of $R_{ij}$
increases smoothly, while under the dHBP it remains roughly constant and then
jumps.  As one varies $a_{ij}$ across its full range, the set
of most likely values for $R_{ij}$ under the dd-IBP spans its full range from
$0$ to $20$, while under the dHBP the most likely value for $R_{ij}$ takes only
a few values.  Instead, varying $a_{ij}$ under the dHBP allows a
variety of bimodal distributions centered near the values from the asymptotic
analysis.

This difference in non-asymptotic behaviors mirrors the difference between the two models in the asymptotic regime, where 
the dd-IBP allows 
feature-sharing-proportions to be specified almost arbitrarily but allows
little flexibility in modeling uncertainty about them, and the dHBP limits
the number of possible values for the feature-sharing proportions, but allows
uncertainty over these values.

\section{Inference using Markov chain Monte Carlo sampling}
\label{sec:mcmc}

Given a dataset $\mathbf{X} = \lbrace \mathbf{x}_i \rbrace_{i=1}^N$ and a latent feature model $P(\mathbf{X}|\mathbf{Z},\theta)$ with parameter $\theta$, the goal of inference is to compute the joint posterior over the
customer assignment matrix $\mathbf{C}$, the dd-IBP hyperparameter
$\alpha$, and likelihood parameter $\theta$, as given by Bayes' rule:
\begin{align}
  P(\mathbf{C},\mathbf{c}^\ast,\theta,\alpha|\mathbf{X},\mathbf{D},f)
  \propto P(\mathbf{X}|\mathbf{C},\mathbf{c}^\ast,\theta) P(\theta)
  P(\mathbf{C}|\mathbf{D},f) P(\mathbf{c}^\ast|\alpha) P(\alpha),
\label{eq:post}
\end{align}
where the first term is the likelihood (recall that $\mathbf{Z}$ is a deterministic function of $\mathbf{C}$ and $\mathbf{c}^\ast$), the second term is the prior over parameters, the third term is the dd-IBP prior over the
connectivity matrix $\mathbf{C}$, the fourth term is the prior over
the ownership vector $\mathbf{c}^\ast$, and the last term is the prior
over $\alpha$. In what follows, we assume that $\mathbf{x}_i$ is conditionally independent of $\mathbf{z}_j$ and $\mathbf{x}_j$ for $j\neq i$ given $\mathbf{z}_i$ and $\theta$.

Exact inference in this model is computationally
intractable. We therefore use MCMC sampling \cite{robert04} to approximate
the posterior with $L$ samples.  The algorithm can be adapted to
different datasets by choosing an appropriate likelihood function. In
the next section, we show how to adapt this algorithm to a simple
linear-Gaussian model.\footnote{Matlab software implementing this algorithm is available at the first author's homepage: \url{www.princeton.edu/~sjgershm}}

Our algorithm combines Gibbs and Metropolis updates. For Gibbs
updates, we sample a variable from its conditional distribution given
the current states of all the other variables.  Conjugacy allows
simple Gibbs updates for $\theta$ and $\alpha$. Because the dd-IBP
prior is not conjugate to the likelihood, we use the Metropolis
algorithm to sample $\mathbf{C}$ and $\mathbf{c}^\ast$.  We generate
proposals for $\mathbf{C}$ and $\mathbf{c}^\ast$, and then accept or
reject them based on the likelihood ratio.  We further divide these
updates into two cases: updates for ``owned'' (active) dishes and
updates of dish ownership.

\textbf{Sampling $\theta$}. To sample the likelihood parameter
$\theta$, we draw from the following conditional distribution:
\begin{align}
P(\theta | \mathbf{X}, \mathbf{C},\mathbf{c}^\ast) \propto P(\mathbf{X}|\mathbf{C},\mathbf{c}^\ast,\theta) P(\theta),
\end{align}
where the prior and likelihood are problem-specific. To
obtain a closed-form expression for this conditional distribution, the
prior and likelihood must be conjugate. For non-conjugate priors, one
can use alternative updates, such as Metropolis-Hastings or slice
sampling \cite{robert04}. Generally, updates for $\theta$ will be
decomposed into separate updates for each component of
$\theta$. In some cases, $\theta$ can be marginalized analytically; an
example is presented in the next section.

\textbf{Sampling $\alpha$}. To sample the hyperparameter $\alpha$, we
draw from the following conditional distribution:
\begin{align}
  P(\alpha | \mathbf{c}^\ast,\mathbf{D},f) \propto P(\alpha)
  \prod_{i=1}^N \mbox{Poisson}(\lambda_i; \alpha/h_i),
\end{align}
where $\lambda_i$ is determined by $\mathbf{c}^\ast$ and the prior on $\alpha$
is a Gamma distribution with shape $\nu_\alpha$ and inverse scale
$\eta_\alpha$. Using the conjugacy of the Gamma and Poisson
distributions, the conditional distribution over $\alpha$ is given by:
\begin{align}
  \alpha | \mathbf{c}^\ast,\mathbf{D},f \sim \mbox{Gamma}\left(\nu_\alpha + \sum_{i=1}^N \lambda_i, \eta_\alpha + \sum_{i=1}^N h_i^{-1}\right).
\end{align}

\textbf{Sampling assignments for owned dishes}. We update customer
assignments for owned dishes (corresponding to ``active'' features)
using Gibbs sampling. For $n=1,\ldots,N$, $i = 1,\ldots,N$, and $k \in \mathcal{K}_n$, we draw a sample from the conditional distribution over
$c_{ik}$ given the current state of all the other variables:
\begin{align}
  P(c_{ik}|\mathbf{c}_{-i},\mathbf{x}_i,\mathbf{c}^\ast,\theta,\mathbf{D},f)
  \propto P(\mathbf{x}_i|\mathbf{C},\mathbf{c}^\ast,\theta)
  P(c_{ik}|\mathbf{D},f),
\label{eq:cond}
\end{align}
where $\mathbf{x}_i$ is the $i$th row of $\mathbf{X}$,
$\mathbf{c}_i$ is the $i$th row of $\mathbf{C}$, and
$\mathbf{c}_{-i}$ is $\mathbf{C}$ excluding row $i$.\footnote{We
  rely on several conditional independencies in this expression; for
  example, $\mathbf{x}_i$ is conditionally independent of
  $\mathbf{X}_{-i}$ given $\mathbf{C},\mathbf{c}^\ast$, and $\theta$.}
The first factor in Eq. \ref{eq:cond} is the likelihood,\footnote{In
  calculating the likelihood, we only include the active columns of
  $\mathbf{Z}$ (i.e., those for which $\sum_{n=1}^N z_{nk} > 0$).} and
the second factor is the prior, given by $P(c_{ik}=j|\mathbf{D},f) =
a_{ij}$. In considering possible assignments of $c_{ik}$, one of two
scenarios will occur: Either data point $i$ reaches the owner of $k$
(in which case feature $k$ becomes active for $i$ as well as for all
other data points that reach $i$), or it does not (in which case feature
$k$ becomes inactive for $i$ as well as for all other data points that
reach $i$). This means we only need to consider two different
likelihoods when updating $c_{ik}$.

\textbf{Sampling dish ownership}. We update dish ownership and
customer assignments for newly owned dishes (corresponding to features
going from inactive to active in the sampling step) using Metropolis
sampling. Both a new connectivity matrix $\mathbf{C}'$ and ownership
vector $\mathbf{c}^{\ast'}$ are proposed by drawing from the prior,
and then accepted or rejected according to a likelihood ratio. In more
detail, the update proceeds as follows.
\begin{enumerate}
\item Propose $\lambda_i' \sim \mbox{Poisson}(\alpha/h_i)$ for each
  data point $i=1,\ldots,N$.
\item Set $\mathbf{C}' \leftarrow \mathbf{C}$. Then populate or
  depopulate it by performing, for each $i=1,\ldots,N$,
  \begin{enumerate}
  \item If $\lambda_i' > \lambda_i$, insert $\lambda_i' -
    \lambda_i$ new dishes into $\mathcal{K}_i$.  Then, for all $k \in [\lambda_i + 1,\lambda_i']$
    and $m \neq i$, sample $c_{mk}'$ according to $P(c_{mk}'=j) =
    a_{mj}$.
  \item If $\lambda_i' < \lambda_i$, remove $\lambda_i' -
    \lambda_i$ randomly selected dishes from $\mathcal{K}_i$.
\end{enumerate}
This reallocation of dishes induces a new ownership vector
$\mathbf{c}^{\ast'}$.
\item Compute the acceptance ratio $\zeta$. Because the prior
  (conditional on the current state of the Markov chain) is being used
  as the proposal distribution, the acceptance ratio reduces to a
  likelihood ratio (the prior and proposal terms cancel out):
\begin{align}
  \zeta =
  \min\left[ 1, \frac{P(\mathbf{X}|\mathbf{C}',\mathbf{c}^{\ast'},\theta)}{P(\mathbf{X}|\mathbf{C},\mathbf{c}^{\ast},\theta)} \right].
\end{align}
\item Draw $r \sim \mbox{Bernoulli}(\zeta)$. Set $\mathbf{C} \leftarrow \mathbf{C}'$ and $\mathbf{c}^\ast \leftarrow \mathbf{c}^{\ast'}$ if $r=1$, otherwise leave $\mathbf{C}$ and $\mathbf{c}^\ast$ unchanged.
\end{enumerate}

Iteratively applying these updates, the sampler will (after a burn-in period) draw samples from a distribution that approaches the posterior (Eq. \ref{eq:post}) as the burn-in period grows large. The time complexity of this algorithm is dominated by the reachability computation, $O(KN^2)$, and the likelihood computation, which is $O(N^3)$ if coded naively (see \cite{griffiths05} for a more efficient implementation using rank-one updates).

\section{A linear-Gaussian model}
\label{sec:lin}

\begin{figure*}
\centering
\includegraphics[width=0.7\textwidth]{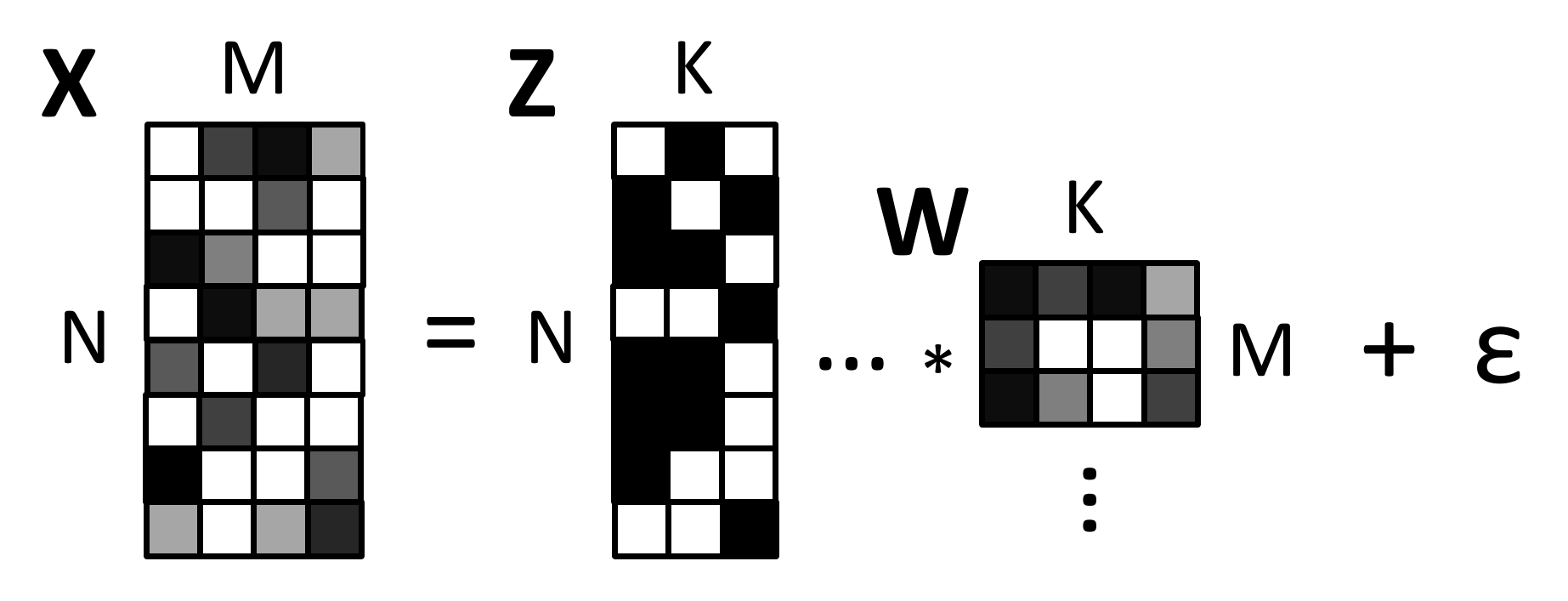}
  	\caption{\textbf{Linear-Gaussian model}. Matrix multiplication view of how latent features ($\mathbf{Z}$) combine with a weight matrix ($\mathbf{W}$) and white noise ($\epsilon$) to produce observed data ($\mathbf{X}$).}
	\label{fig:schematic_linear}
\end{figure*}

As an example of how the dd-IBP can be used in data analysis, we
incorporate it into a linear-Gaussian latent feature model (Figure
\ref{fig:schematic_linear}). This model was originally studied for the
IBP by Griffiths and Ghahramani \cite{griffiths05,griffiths11}. The observed data $\mathbf{X}
\in \mathbb{R}^{N \times M}$ consist of $N$ objects, each of which is
a $M$-dimensional vector of real-valued object properties. We model
$\mathbf{X}$ as a linear combination of binary latent features
corrupted by Gaussian noise:
\begin{align}
\mathbf{X} = \mathbf{Z}\mathbf{W} + \epsilon,
\end{align}
where $\mathbf{W}$ is a $K \times M$ matrix of real-valued weights,
and $\epsilon$ is a $N \times M$ matrix of independent, zero-mean
Gaussian noise terms with standard deviation $\sigma_x$. We place a
zero-mean Gaussian prior on $\mathbf{W}$ with covariance $\sigma_w^2
\mathbf{I}$. Intuitively, the weights capture how the latent features
interact to produce the observed data. For example, if each latent
feature corresponds to a person in an image, then the weight $w_{km}$
captures the contribution of person $k$ to pixel $m$.

Within the algorithm of the previous section, $\theta =
\mathbf{W}$. As a consequence of our Gaussian assumptions,
$\mathbf{W}$ can be marginalized analytically, yielding the
likelihood:
\begin{align}
  P(\mathbf{X}|\mathbf{Z}) &= \int_{\mathbf{W}} P(\mathbf{X}|\mathbf{Z},\mathbf{W}) P(\mathbf{W}) d\mathbf{W} \nonumber \\
  &= \frac{\exp\left\{ - \frac{1}{2 \sigma_x^2} \mbox{tr}\left(\mathbf{X}^T(\mathbf{I}-\mathbf{Z}\mathbf{H}^{-1}\mathbf{Z}^T)\mathbf{X} \right) \right\}}
{(2\pi)^{NM/2} \sigma_x^{(N-K)M}\sigma_w^{KM} |\mathbf{H}|^{M/2}},
\label{eq:lik}
\end{align}
where tr($\cdot$) is the matrix trace and $\mathbf{H} =
\mathbf{Z}^T\mathbf{Z} + \frac{\sigma^2_x}{\sigma^2_w} \mathbf{I}$. In
calculating the likelihood, we only include the ``active'' columns of
$\mathbf{Z}$ (i.e., those for which $\sum_{j=1}^N z_{jk} > 0$), and
$K$ is the number of active columns.

\section{Experimental results}
\label{sec:results}

In this section we report experimental investigations of the dd-IBP
and comparisons with alternative models. We first show how the dd-IBP
can be used as a dimensionality reduction pre-processing technique for
classification tasks when the data points are
non-exchangeable. We then show how the dd-IBP can be applied to
missing data problems.

\subsection{Dimensionality reduction for classification}

The performance of supervised learning algorithms is often enhanced by
pre-processing the data to reduce its dimensionality
\cite{bishop06}. Classical techniques for dimensionality reduction,
such as principal components analysis and factor analysis, assume
exchangeability, as does the infinite latent feature model based on the IBP \cite{griffiths05}. For this
reason, these techniques may not work as well for pre-processing
non-exchangeable data, and this may adversely affect their performance
on supervised learning tasks.

We investigated this hypothesis using a magnetic resonance imaging
(MRI) data set collected from 27 patients with Alzheimer's disease and
35 healthy controls \cite{christou11}.\footnote{Available at:
  \url{http://wiki.stat.ucla.edu/socr/index.php/SOCR_Data_July2009_ID_NI}.}
The observed features consist of 4 structural summary statistics
measured in 56 brain regions of interest: (1) surface area; (2) shape
index; (3) curvedness; (4) fractal dimension. The classification task
is to sort individuals into Alzheimer's or control classes based on
their observed features.

Age-related changes in brain structure produce natural declines in
cognitive function that make diagnosis of Alzheimer's disease
difficult \cite{erkinjuntti86}. Thus, it is important to
take age into account when designing predictive models. For the dd-IBP
and dHBP, age is naturally incorporated as a covariate over which we
constructed a distance matrix.  Specifically, we defined $d_{ij}$ as
the absolute age difference between subjects $i$ and $j$, with $d_{ij}=\infty$ for $j>i$ (i.e., the distance matrix is sequential). This induces
a prior belief that individuals with similar ages tend to share more
latent features. In the MRI data set, ages ranged from 60 to 90
(median: $76.5$).

\begin{figure*}
\centering
\includegraphics[width=0.6\textwidth]{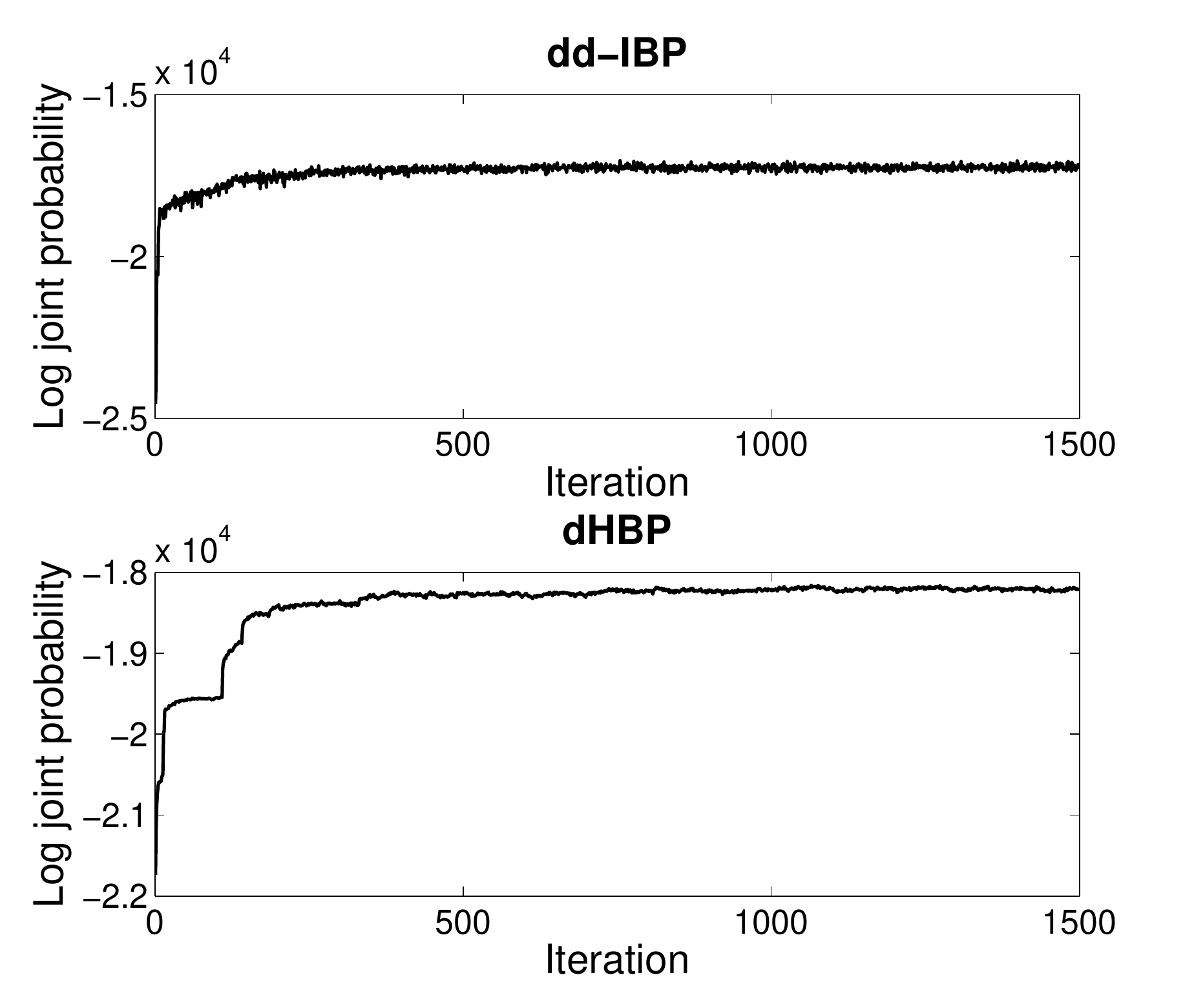}
  	\caption{\textbf{Trace plots}. Representative traces of the log joint probability of the Alzheimer's data and latent variables for the dd-IBP (\emph{top}) and dHBP (\emph{bottom}). Each iteration corresponds to a sweep over all the latent variables.}
	\label{fig:ad_trace}
\end{figure*}

In detail, we ran 1500 iterations of MCMC sampling on the entire data set using the linear-Gaussian observation model, and then selected the latent features of the \emph{maximum a posteriori} sample as input to a supervised learning algorithm ($L_2$-regularized logistic regression, with the regularization constant set to $10^{-6}$). Training was performed on half of the data, and testing on the other half.\footnote{A few randomly chosen individuals were removed from the test set to make it balanced.} The noise hyperparameters of the dd-IBP and dHBP ($\sigma_x$ and $\sigma_y$) were updated using Metropolis-Hastings proposals.

We monitored the log of the joint distribution $\log P(\mathbf{X},\alpha,\mathbf{C},\mathbf{c}^\ast)$. Visual inspection of the log joint probability traces suggested that the sampler reaches a local maximum within 400-500 iterations (Figure \ref{fig:ad_trace}). This process was repeated for a range of decay parameter ($\beta$) values, using the exponential decay function.  The same proximity matrix, $\mathbf{A}$, was used for both the dd-IBP and dHBP. We performed 5 random restarts of the sampler and recomputed the classification measure for each restart, averaging the resulting measures to reduce sampling variability. For comparison, we also made predictions using the standard IBP, the dIBP \cite{williamson10}, and the raw observed features (i.e., no pre-processing). The dIBP was fit using the MCMC algorithm described in Williamson et al. \cite{williamson10}, which adaptively samples the parameters controlling dependencies between observations (thus the results do not depend on $\beta$).

\begin{figure*}
\centering
\includegraphics[width=0.6\textwidth]{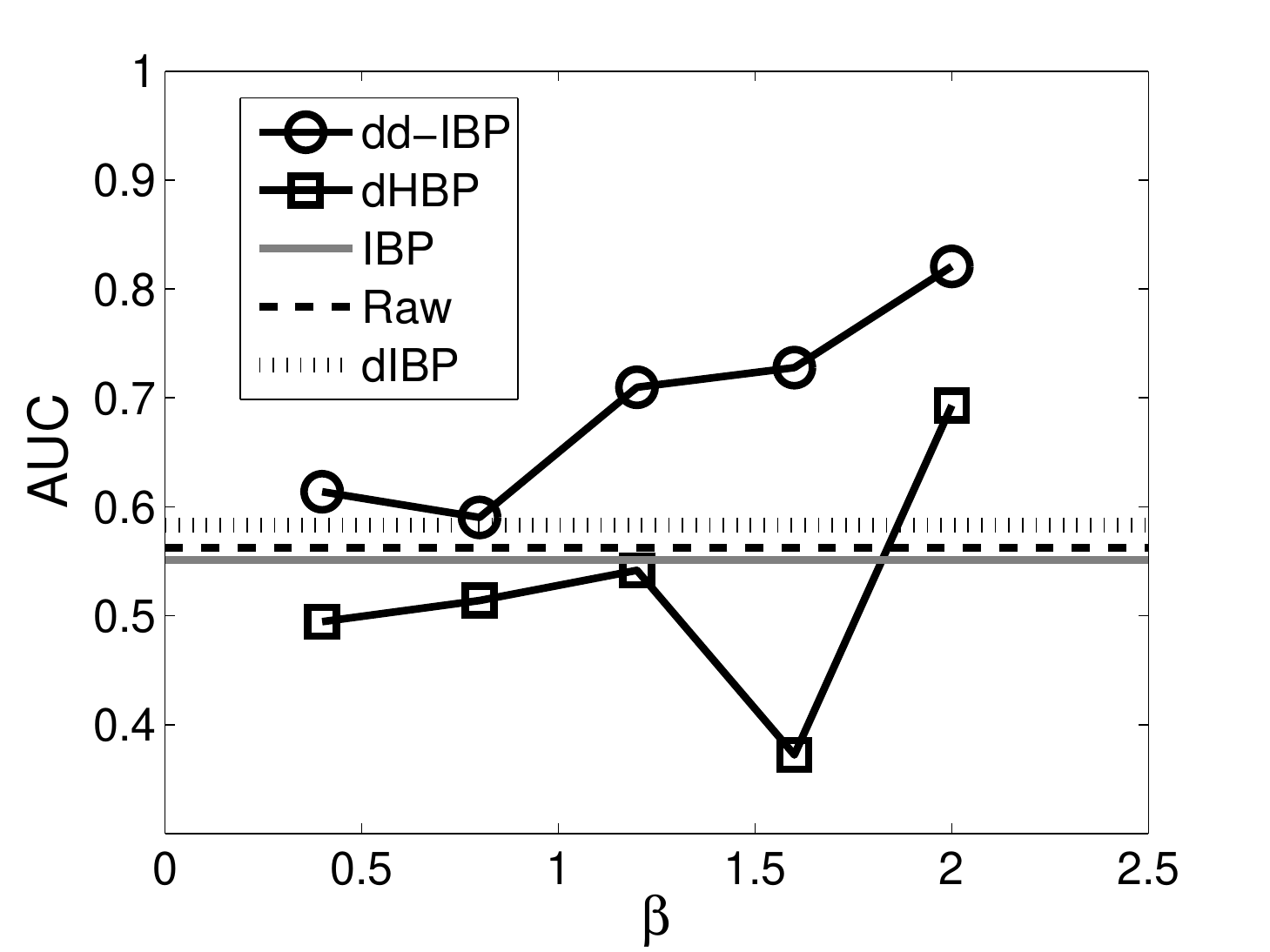}
  	\caption{\textbf{Classification results for Alzheimer's data set}. Area under the curve (AUC) for binary classification (Alzheimer's vs. normal control) using $L_2$-regularized logistic regression and features learned from a linear-Gaussian latent feature model. Each curve represents a different choice of predictor variables (latent features) for logistic regression. The x-axis corresponds to different settings of the exponential decay function parameter, $\beta$. ``Raw'' refers to the original data features (see text for details); the IBP, dd-IBP, dIBP and dHBP results were based on using the latent features of the \emph{maximum a posteriori} sample following 1500 iterations of MCMC sampling.}
	\label{fig:ad_results}
\end{figure*}

Classification results are shown in Figure \ref{fig:ad_results} (left), where performance is measured as the area under the receiver operating characteristic curve (AUC). Chance performance corresponds to an AUC of 0.5, perfect performance to an AUC of 1. For a range of $\beta$ values, the dd-IBP produces superior classification performance to the alternative models, with performance increasing as a function of $\beta$. The dHBP performs worse relative to the raw data for low $\beta$ values. The magnitude of the standard error (across random restarts) is roughly $1/10$ that of the means.

We also ran the dd-IBP sampler with $\beta=0$ (in which case the dd-IBP and IBP are equivalent) and found no significant difference between it and the standard IBP sampler with respect to performance on the Alzheimer's classification and the EEG reconstruction (see next section).

\subsection{Reconstructing missing data}

As an example of a missing data problem, we use latent feature models to reconstruct missing observations in electroencephalography (EEG) time series. The EEG data\footnote{Available at: \url{http://mmspl.epfl.ch/page33712.html}} are from a visual detection experiment in which human subjects were asked to count how many times a particular image appeared on the screen \cite{hoffmann08}. The data were collected as part of a larger effort to design brain-computer interfaces to assist physically disabled subjects.

\begin{figure*}
\centering
\includegraphics[width=0.6\textwidth]{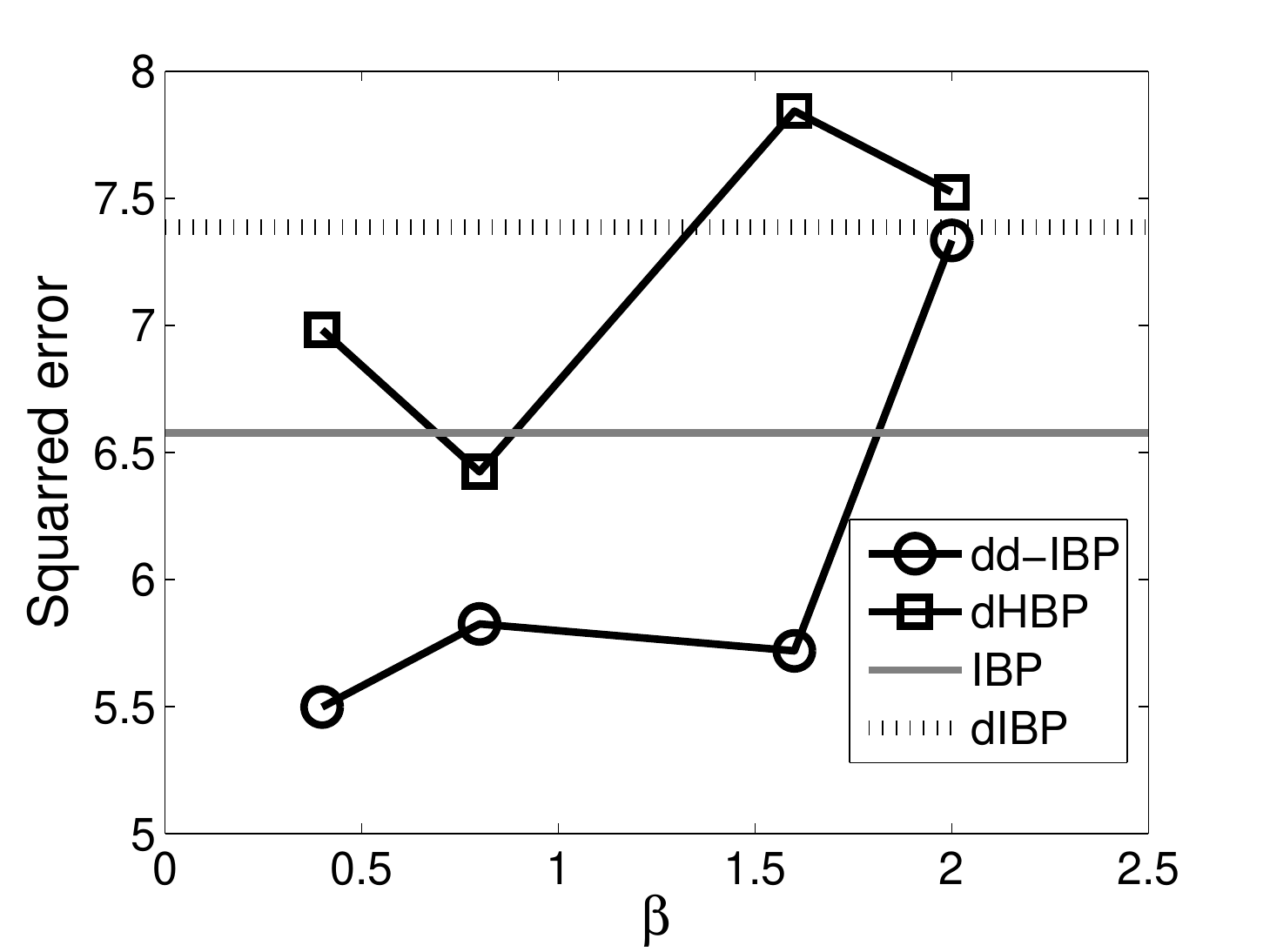}
  	\caption{\textbf{Reconstruction of missing EEG data}. Reconstruction error for latent feature models as a function of the exponential decay function parameter, $\beta$. Results were based on the \emph{maximum a posteriori} sample following 1500 iterations of MCMC sampling. Lower values indicate better performance.}
	\label{fig:eeg}
\end{figure*}

Distance between data points was defined using the absolute time-difference. Data were z-scored prior to analysis. For 10 of the data points, we removed 2 of the observed features at random. We then ran the MCMC sampler for 1500 iterations, adding Gibbs updates for the missing data by sampling from the observation distribution (Eq. \ref{eq:lik}) conditional on the current values of the latent features and hyperparameters. We then used the MAP sample for reconstruction. We measured performance by the squared reconstruction error on the missing data. Figure \ref{fig:eeg} shows the reconstruction results, demonstrating that the dd-IBP is effective for reconstructing missing data in this dataset and achieves lower reconstruction error than the alternative models we consider.

\section{Conclusions}

By relaxing the exchangeability assumption for infinite latent feature models, the dd-IBP extends their applicability to a richer class of data. We have shown empirically that this innovation fares better than the standard IBP on non-exchangeable data (e.g., timeseries).

We note that the dd-IBP is not a standard Bayesian nonparametric distribution, in the sense of arising from a de Finetti mixing distribution. For the standard IBP, the de Finetti mixing distribution has been identified as the beta process \cite{thibaux07}, but this result does not generalize to the dd-IBP due to its non-exchangeability (a consequence of de Finetti's theorem). Nonetheless, this does not detract from our model's ability to let the data infer the number of latent features, a property that it shares with other infinite latent feature models.

A number of future directions are possible. First, it may be possible to exploit distance dependence to derive more efficient samplers. In particular, Doshi-Velez and Ghahramani \cite{doshi09a} have shown that partitioning the data into subsets enables faster Gibbs sampling for the traditional IBP; the window decay function imposes a natural partition of the data into conditionally independent subsets.

Second, application of the dd-IBP to other likelihood functions is straightforward. For example, it could be applied to relational data \cite{meeds07,miller09} or text data \cite{thibaux07}. As pointed out by Miller et al. \cite{miller09}, covariates like age or location often play an important role in link prediction. Whereas Miller et al. \cite{miller09} incorporated covariates into the likelihood function, one could instead incorporate them into the prior by defining covariate-based distances between data points (e.g., the age difference between two people). A distinction of the latter approach is that it would allow one to model dependencies in terms of latent features. For instance, two people close in age or geographic location may be more likely to share latent interests, a pattern naturally captured by the dd-IBP.

Third, modeling shared dependency structure across groups is important for several applications. In fMRI and EEG studies, for example, similar spatial and temporal dependencies are frequently observed across subjects. Modeling shared structure without sacrificing intersubject variability has been addressed with hierarchical models \cite{beckmann03,woolrich04}. One way to extend the dd-IBP hierarchically would be to allow the parameters of the decay function to vary across individuals while being coupled together by higher-level variables.

\subsubsection*{Acknowledgements}

SJG was supported by a NSF graduate research fellowship. PIF acknowledges support from NSF Award $\#142251$. We thank Matt Hoffman, Chong Wang, Gungor Polatkan, Sean Gerrish and John Paisley for helpful discussions. We are also grateful to Sinead Williamson and Mingyuan Zhou for sharing their code.

\bibliographystyle{plain}

\section*{Appendix: proofs}

Recall that $R_i = \sum_{k=1}^\infty z_{ik}$ is the number of features held by data point $i$, and $R_{ij}=\sum_{k=1}^\infty z_{ik}z_{jk}$ is the number of features shared by data points $i$ and $j$, where $i\ne j$.

\subsection*{Proposition 1}

\textit{
Under the dd-IBP,
\begin{align}
R_i &\sim \mathrm{Poisson}\left( \alpha \sum_{n=1}^N h_n^{-1} P(\mathcal{L}_{in}=1) \right), \\
R_{ij} &\sim \mathrm{Poisson} \left( \alpha \sum_{n=1}^N h_n^{-1} P(\mathcal{L}_{in}=1,\mathcal{L}_{jn}=1) \right).
\end{align}
}

\begin{proof}
For each feature, there is some probability $\pi_i$ that it is turned on for data point $i$, and some probability $\pi_{ij}$ that it is shared by $i$ and $j$ (note that features are exchangeable in the dd-IBP). The total number of features across all data points is distributed according to $\mbox{Poisson}(\lambda)$, where $\lambda = \alpha \sum_{n=1}^N h_n^{-1}$. Since each feature is turned on independently, the total number of active features for a single data point $i$ is distributed according to $R_i \sim \mbox{Poisson}(\lambda \pi_i)$. Similarly, the total number of features shared by data points $i$ and $j$ is $R_{ij} \sim \mbox{Poisson}(\lambda \pi_{ij})$. The activation and co-activation probabilities are given by:
\begin{align}
\pi_i &= \sum_{n=1}^N P(c^*=n) P(\mathcal{L}_{in}=1) = \frac{ \sum_{n=1}^N h_n^{-1} P(\mathcal{L}_{in}=1)}{\sum_{j=1}^N h_j^{-1}}, \\
\pi_{ij} &= \sum_{n=1}^N P(c^*=n) P(\mathcal{L}_{in}=1,\mathcal{L}_{jn}=1) = \frac{ \sum_{n=1}^N h_n^{-1} P(\mathcal{L}_{in}=1,\mathcal{L}_{jn}=1)}{\sum_{j=1}^N h_j^{-1}},
\end{align}
where we have used that $P(c^*=n) = h_n^{-1} / \sum_{j=1}^N h_j^{-1}$.
\end{proof}

\subsection*{Proposition 2}

\textit{
If $B_0$ is continuous, then under the dHBP,
\begin{align}
  R_i|\mathbf{g}_{1:N} &\sim \mathrm{Poisson}\left(\gamma\right), \\
  R_{ij}|\mathbf{g}_{1:N} &\sim \left\{ \begin{array}{ll}
      \mathrm{Poisson}\left(\gamma\frac{c_0 + c_1 + 1}{(c_0+1)(c_1+1)} \right) & \mbox{if $g_i = g_j$},\\
      \mathrm{Poisson}\left(\gamma\frac{1}{c_0+1} \right) & \mbox{if
        $g_i \neq g_j$.}\end{array} \right.
\end{align}
}

\begin{proof}

  We write the random measures $B$ and $B_j^*$ in the generative model defining the dHBP in Section 3.2 as the following mixtures over point masses.
  \begin{align}
    B &= \sum_{k=1}^\infty p_k \delta_{\omega_k},\qquad 
    p_{k} \sim \mbox{Beta}(0, c_0), \quad
    \omega_k \sim B_0. \\
    B^\ast_j &= \sum_{k=1}^\infty p_{jk}^\ast \delta_{\omega_k}, \qquad
    p^\ast_{jk} \sim \mbox{Beta}(c_1 p_k, c_1(1-p_k)).
  \end{align}
  Recall that $X_i \sim \mbox{BeP}(B^*_{g_i})$ where $g_i \sim \mbox{Multinomial}(\mathbf{a}_i)$.


  Let $z_{ik}$ be the random variable that is $1$ if the Bernoulli process draw $X_i$ has atom $\omega_k$, and $0$ if not.  We have $z_{ik} \sim \mbox{Bernoulli}(p^\ast_{g_ik})$.
Because $B_0$ is continuous, $P(\omega_k = \omega_{k'})=0$ for $k\ne k'$ and 
the random variables $R_i$ and $R_{ij}$ satisfy
\begin{align}
R_i = \sum_{k=1}^K z_{ik} \quad \mbox{and} \quad R_{ij} = \sum_{k=1}^K z_{ik}z_{jk}.
\end{align}
We first show that $R_i$ is Poisson distributed with mean $\gamma$.

Let $q_i(\epsilon)$ denote the probability that $X_i$ has atom $\omega_k$ conditioned on $p_k>\epsilon$ (this value does not depend on $k$).  That is,
  $q_i(\epsilon) = P(z_{ik}=1 | p_k>\epsilon)$.

For a given $\epsilon$, the density of $p_k$ conditioned on $p_k > \epsilon$ is:
\begin{align}
P(p_k\in dp|p_k>\epsilon) = \frac{c_0 p^{-1}(1-p)^{c_0-1}}{\int_{\epsilon}^1 c_0 u^{-1}(1-u)^{c_0-1} du} dp, \quad p \in (\epsilon,1).
\end{align}
We can use this density to calculate the success probability $q_i(\epsilon)$:
\begin{equation}
  q_i(\epsilon) = \mathbb{E}[z_{ik}|p_k > \epsilon] 
  = \mathbb{E}[p^\ast_{g_ik}|p_k > \epsilon] 
= \mathbb{E}[p_k|p_k > \epsilon] 
= \frac{\int_{\epsilon}^1 p c_0 p^{-1}(1-p)^{c_0-1} dp}{\int_{\epsilon}^1 c_0 p^{-1}(1-p)^{c_0-1} dp},
\end{equation}
where we have used the tower property of conditional expectation in the second and third equalities.

For a given $\epsilon>0$, let $N_\epsilon$ denote the number of atoms in $B$ with $p_k > \epsilon$. This number is Poisson-distributed with mean $\lambda_\epsilon = \gamma \int_\epsilon^1 c_0 p^{-1} (1-p)^{c_0-1} dp$.  

Let $R_i(\epsilon)$ be the number of such atoms that are also in $X_i$.
Because $R_i(\epsilon)$ is the sum of $N_\epsilon$ independent Bernoulli trials that each have success probability $q_i(\epsilon)$, it follows that
$R_i(\epsilon)|N_\epsilon \sim \mbox{Binomial}(N_\epsilon,q_i(\epsilon))$ and
\begin{equation}
  R_i(\epsilon) \sim \mathrm{Poisson}(\lambda_\epsilon q_i(\epsilon)).
\end{equation}
Because $R_i = \lim_{\epsilon\to0} R_i(\epsilon)$, 
it follows that
$R_i \sim \mbox{Poisson}(\lim_{\epsilon \rightarrow 0} \lambda_\epsilon q_i(\epsilon))$, where
\begin{align}
\lim_{\epsilon \rightarrow 0} \lambda_\epsilon q_i(\epsilon) &= 
\lim_{\epsilon \rightarrow 0} \left[\gamma \int_{\epsilon}^1 c_0 p^{-1}(1-p)^{c_0-1} dp\right] \left[\frac{\int_{\epsilon}^1 p c_0 p^{-1}(1-p)^{c_0-1} dp}{\int_{\epsilon}^1 c_0 p^{-1}(1-p)^{c_0-1} dp} \right] \nonumber \\
&= \lim_{\epsilon \rightarrow 0} \gamma \int_{\epsilon}^1 p c_0 p^{-1}(1-p)^{c_0-1} dp \nonumber 
= \gamma c_0 \int_{0}^1 (1-p)^{c_0-1} dp \nonumber 
= \gamma,
\end{align}
where we have used that 
$\int_{0}^1 (1-p)^{c_0-1} dp = \frac1{c_0}$.
Thus $R_i \sim \mbox{Poisson}(\gamma)$.

We perform a similar analysis to show the distribution of $R_{ij}$. 
Let $q_{ij}(\epsilon)$ denote the probability that $X_i$ and $X_j$ share atom $\omega_k$ conditional on $p_k>\epsilon$, $g_i$ and $g_j$.  That is, 
\begin{equation}
  q_{ij}(\epsilon) = P(z_{ik}=z_{ij}=1 | g_i, g_j, p_k>\epsilon).
\end{equation}
Although only $\epsilon$ appears in the argument of $q_{ij}(\epsilon)$, this quantity also implicitly depends on $g_i$ and $g_j$.  We calculate $q_{ij}(\epsilon)$ explicitly below.

Let $R_{ij}(\epsilon)$ be the number of atoms $\omega_k$ for which $p_k>\epsilon$ and $\omega_k$ is in both $X_i$ and $X_j$.
We have $R_{ij}(\epsilon)|N_\epsilon,g_i,g_j \sim \mbox{Binomial}(N_\epsilon,q_{ij}(\epsilon))$ and 
\begin{align}
R_{ij}(\epsilon)|g_i,g_j \sim \mbox{Poisson}(\lambda_\epsilon q_{ij}(\epsilon)).
\end{align}
Because $R_{ij} = \lim_{\epsilon\to0} R_{ij}(\epsilon)$, 
it follows that 
$R_{ij} \sim \mbox{Poisson}(\lim_{\epsilon \rightarrow 0} \lambda_\epsilon q_{ij}(\epsilon))$.

To calculate $\lim_{\epsilon \rightarrow 0} \lambda_\epsilon q_{ij}(\epsilon)$,
we consider two cases.  In each case, we first calculate $q_{ij}(\epsilon)$ and then calculate the limit, showing that it is the same as the mean of $R_{ij}$ claimed in the statement of the proposition.
\begin{itemize}
\item \textbf{Case 1}: $g_i = g_j$
\begin{align}
q_{ij}(\epsilon) &= \mathbb{E}[z_{ik} z_{jk}|p_k > \epsilon,g_i,g_j] \nonumber 
= \mathbb{E}[(p^\ast_{g_ik})^2|p_k > \epsilon,g_i,g_j] \nonumber \\
&= \mathbb{E}[\mathbb{E}[(p^\ast_{g_ik})^2|p_k,g_i,g_j]|p_k > \epsilon,g_i,g_j] \nonumber 
= \mathbb{E}[p_k (c_1 p_k + 1)/(c_1+1)|p_k > \epsilon,g_i,g_j] \nonumber \\
&= \frac{\int_{\epsilon}^1 \frac{c_1 p + 1}{c_1+1} p c_0 p^{-1}(1-p)^{c_0-1} dp}{\int_{\epsilon}^1 c_0 p^{-1}(1-p)^{c_0-1} dp} 
= \gamma \frac{c_0}{c_1+1} \frac{\int_{\epsilon}^1 (c_1p+1) (1-p)^{c_0-1} dp}{\lambda_\epsilon}.
\end{align}
Then the limit $\lim_{\epsilon\to0} \lambda_\epsilon q_{ij}(\epsilon)$ can be written
\begin{align}
\lim_{\epsilon \rightarrow 0} \lambda_\epsilon q_{ij}(\epsilon) 
&= \gamma \frac{c_0}{c_1+1} \int_0^1 (c_1p+1) (1-p)^{c_0-1} dp \nonumber \\
&= \gamma \frac{c_0}{c_1+1} \left[ c_1 \int_0^1 p (1-p)^{c_0-1} dp + \int_0^1 (1-p)^{c_0-1} dp\right] \nonumber  \\
&= \gamma \frac{c_0}{c_1+1} \left[ \frac{c_1}{c_0(c_0+1)} + \frac{1}{c_0} \right] \nonumber 
= \gamma \frac{c_0 + c_1 + 1}{(c_0+1)(c_1+1)},
\end{align}
where we have used that 
$\int_{0}^1 (1-p)^{c_0-1} dp = \frac1{c_0}$ and 
$\int_{0}^1 p(1-p)^{c_0-1} dp = \frac1{c_0(c_0+1)}$.

\item \textbf{Case 2}: $g_i \neq g_j$
\begin{align}
q_{ij}(\epsilon) 
&= \mathbb{E}[z_{ik} z_{jk}|p_k > \epsilon,g_i,g_j] \nonumber 
= \mathbb{E}[\mathbb{E}[p^\ast_{g_ik} p^\ast_{g_jk} |p_k,g_i,g_j]|p_k > \epsilon,g_i,g_j] \nonumber \\
&= \mathbb{E}[p_k^2|p_k > \epsilon,g_i,g_j] \nonumber 
= \gamma \frac{\int_{\epsilon}^1 p^2 c_0 p^{-1}(1-p)^{c_0-1} dp}{\lambda_\epsilon}.
\end{align}
Then the limit $\lim_{\epsilon\to0} \lambda_\epsilon q_{ij}(\epsilon)$ can be written
\begin{align}
\lim_{\epsilon \rightarrow 0} \lambda_\epsilon q_{ij}(\epsilon) 
= \gamma c_0 \int_0^1 p(1-p)^{c_0-1} dp \nonumber  
= \gamma \frac{1}{c_0+1},
\end{align}
where we have used that 
$\int_{0}^1 p(1-p)^{c_0-1} dp = \frac1{c_0(c_0+1)}$.

\end{itemize}
\end{proof}

\end{document}